%% file: main.tex
\documentclass{article} 
\usepackage{iclr2025_conference,times}

\input{math_commands.tex}

\usepackage{graphicx}
\usepackage{hyperref}
\usepackage{enumerate}
\usepackage{url}
\usepackage{algorithm}
\usepackage{algpseudocode}
\usepackage{multicol}
\usepackage{ulem} 
\usepackage{subcaption}
\usepackage{booktabs} 
\usepackage{arydshln}
\usepackage{tocloft}
\usepackage{titletoc}

\newtheorem{theorem}{Theorem}
\newtheorem{proofpart}{Proof}

\title{Rectified Diffusion: Straightness Is Not Your Need in Rectified Flow}

\author{
  Fu-Yun Wang\textsuperscript{1} \quad Ling Yang\textsuperscript{2} \quad Zhaoyang Huang\textsuperscript{1} \\[1ex]
  \textbf{Mengdi Wang\textsuperscript{3} \quad Hongsheng Li\textsuperscript{1}}\\[2ex]
  \textsuperscript{1}MMLab, CUHK, Hong Kong SAR\\[1ex]
  \textsuperscript{2}Peking University, Beijing, China\\[1ex]
\textsuperscript{3}Princeton University, New Jersey, USA\\[2ex]
  \texttt{fywang@link.cuhk.edu.hk} \quad \texttt{yangling0818@163.com} \\ 
  \texttt{drinkingcoder@link.cuhk.edu.hk} \\
   \texttt{mengdiw@princeton.edu} \\
   \texttt{hsli@ee.cuhk.edu.hk} 
}

\iclrfinalcopy 
\begin{document}

\maketitle

\begin{abstract}
Diffusion models have greatly improved visual generation but are hindered by slow generation speed due to the computationally intensive nature of solving generative ODEs. Rectified flow, a widely recognized solution, improves generation speed by straightening the ODE path. Its 
key components include: 1) using the diffusion form of flow-matching, 2) employing $\boldsymbol v$-prediction, and 3) performing rectification (a.k.a. reflow). In this paper, we argue that the success of rectification primarily lies in using a pretrained diffusion model to obtain matched pairs of noise and samples, followed by retraining with these matched noise-sample pairs. Based on this, components 1) and 2) are unnecessary. Furthermore, we highlight that straightness is not an essential training target for rectification; rather, it is a specific case of flow-matching models. The more critical training target is to achieve a first-order approximate ODE path, which is inherently curved for models like DDPM and Sub-VP. Building on this insight, we propose Rectified Diffusion, which generalizes the design space and application scope of rectification to encompass the broader category of diffusion models, rather than being restricted to flow-matching models. We validate our method on Stable Diffusion v1-5 and Stable Diffusion XL. Our method not only greatly simplifies the training procedure of rectified flow-based previous works~(e.g., InstaFlow) but also achieves superior performance with even lower training cost. Our code is available at \url{https://github.com/G-U-N/Rectified-Diffusion}.
\end{abstract}

\input{secs/sec1_intro}

\input{secs/sec3_analysis}

\input{secs/sec5_exp}

\clearpage

\bibliography{main}
\bibliographystyle{iclr2025_conference}

\appendix

\clearpage

\setcounter{page}{1}
\section*{\Huge Appendix}

\startcontents[appendices]
\printcontents[appendices]{l}{1}{\setcounter{tocdepth}{2}}

\renewcommand{\thesection}{\Roman{section}}

\input{secs/sec2_related_work}

\input{secs/sec_theorem}

\end{document}

%% file: math_commands.tex
\usepackage{amsmath,amsfonts,bm}

\def\eqref#1{equation~\ref{#1}}

\def\1{\bm{1}}

\DeclareMathAlphabet{\mathsfit}{\encodingdefault}{\sfdefault}{m}{sl}
\SetMathAlphabet{\mathsfit}{bold}{\encodingdefault}{\sfdefault}{bx}{n}



%% file: secs/sec1_intro.tex
\section{Introduction}

Diffusion models have greatly advanced the field of visual generation, enabling the creation of high-quality images and vivid videos from text~\citep{ddpm,sde,rombach2022high,makeavideo,sdxl,sd3,shi2024motion}.  However, the generation process of diffusion models involves solving an expensive generative ODE numerically, which significantly slows down the generation speed compared to other generative models~(e.g., GAN)~\citep{gan,stylegant,sauer2023stylegan}.  A widely recognized solution to this issue is rectified flow. The training target of rectified flow, as highlighted in the previous works~\citep{instaflow,rectifiedflow,yan2024perflow}, is to make the new ODE path straighter, enabling the models to generate high-fidelity images with fewer steps while retaining the flexibility of sampling with more inference steps for further quality enhancement.  The key components of rectified flow are threefold: 
\begin{enumerate}[\arabic{enumi})]
    \item \textbf{Flow-Matching.} Rectified flow proposes to employ the flow-matching based diffusion form~\citep{rectifiedflow,flowmathcing}. The intermediate noisy state $\mathbf x_t$ is defined as $ (1-t)\mathbf x_0 + t \boldsymbol{\epsilon}$, where $\mathbf x_0$ is the clean data, $\boldsymbol \epsilon\sim \mathcal N(\boldsymbol{0}, \mathbf I)$ is normal noise, and $t \in [0,1]$ is the timestep. This design is more straightforward compared to the semi-linear form of the original DDPM~\citep{ddpm}.
    \item \textbf{$\boldsymbol v$-prediction.} Rectified flow proposes to adopt $\boldsymbol v$-prediction~\citep{progressivedistillation,rectifiedflow}. That is, the model learns to predict $\boldsymbol v = \mathbf x_0 - \boldsymbol{\epsilon}$. This makes the denoising form simple. For example, one can predict $\mathbf x_0$ based on $\mathbf x_t$ with $\hat{\mathbf x}_0 = \mathbf x_t + t \boldsymbol \hat{v}_{\boldsymbol \theta}$, where $\boldsymbol{\theta}$ denotes the model parameters and $\hat{\phantom{x}}$ denotes the predictions. Moreover, it avoids the numerical issue when $t \approx 1$ with $\boldsymbol \epsilon$-prediction. For example, $\hat{\mathbf x}_0 = \frac{\mathbf x_t - t \hat{\boldsymbol \epsilon}_{\boldsymbol \theta}}{1-t} \approx \frac{\mathbf x_t - t \hat{\boldsymbol \epsilon}_{\boldsymbol \theta}}{0}$, which is invalid.  
    \item \textbf{Rectification.} Rectification, also known as reflow, is an important technique proposed in  rectified flow~\citep{rectifiedflow}. It is a progressive retraining technique that greatly improves the generation quality at low-step regime and maintains the flexibility of standard diffusion models. To be specific, it turns an arbitrary coupling of $\mathbf x_0 \sim \mathbb P_0$~(real data) and $\boldsymbol \epsilon \sim \mathbb P_1$~(noise) adopted in standard diffusion training to a new deterministic coupling of $\hat{\mathbf x}_0 \sim \mathbb P^{\boldsymbol \theta}_0$~(generated data) and $\boldsymbol \epsilon \sim \mathbb P_1$~(pre-collected noise). To put it in a nutshell, it replaces the $\mathbf x_t = (1-t) \mathbf x_0 + t \boldsymbol \epsilon$ with $\mathbf x_t = (1-t) \hat{\mathbf x}_0 + t \hat{\boldsymbol \epsilon}$, where $\mathbf x_0$ is real data, $\hat{\mathbf x}_0$ is data generated by pretrained diffusion models $\boldsymbol \theta$, $\boldsymbol \epsilon$ is the randomly sampled noise, and $\hat{\boldsymbol \epsilon}$ is the noise used to generate data $\hat{\mathbf x}_0$. Previous works emphasize the rectification procedure is only feasible to $\boldsymbol v$-prediction based flow-matching models.  That is, they believe the first two points are the foundations to adopt rectification for improving efficiency. And they emphasize the rectification procedure `straightens' the ODE path. 
\end{enumerate}

\begin{figure}[t]
    \centering
\includegraphics[width=0.95\linewidth]{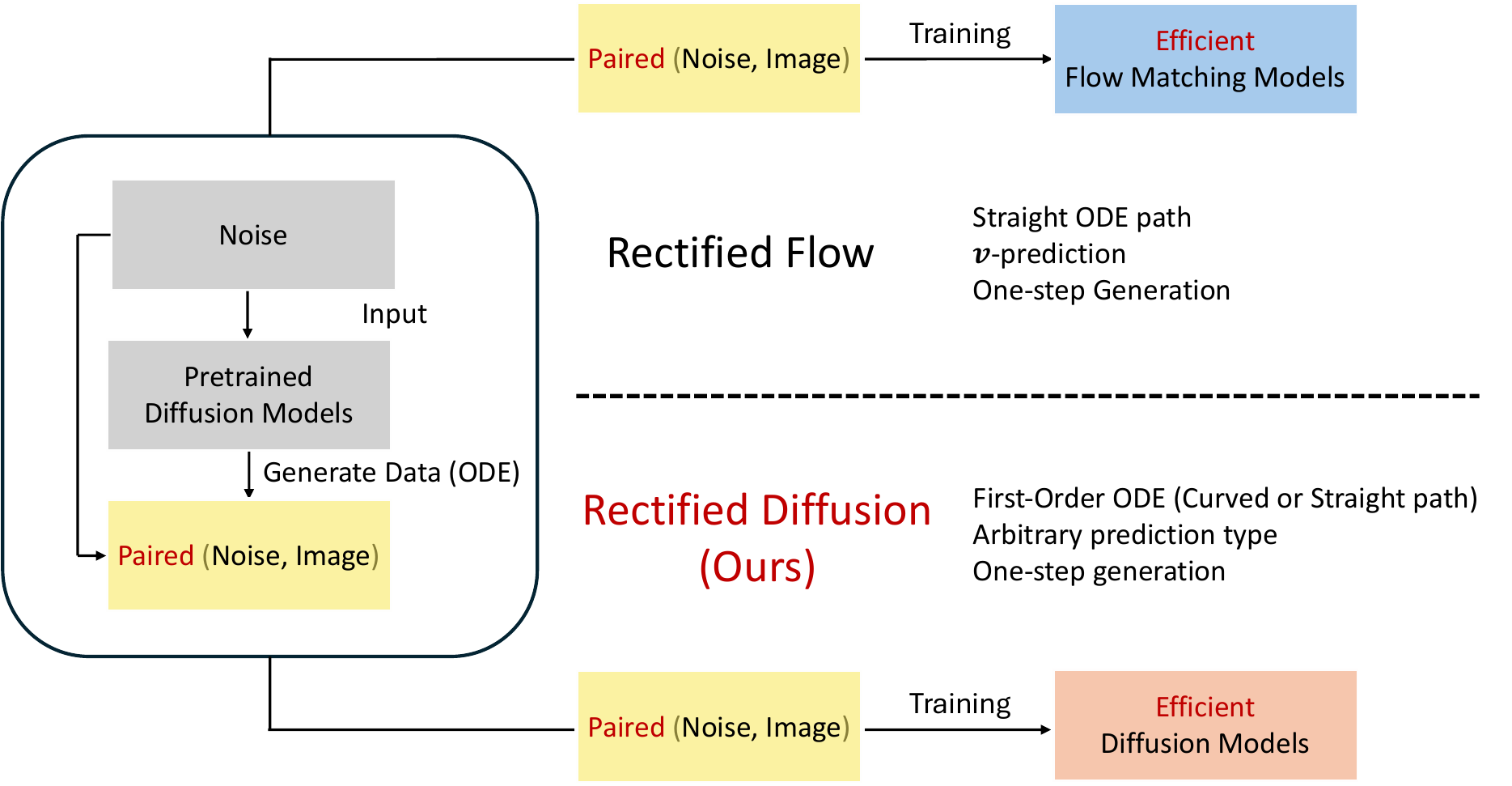}
    \caption{Overview of comparison between   Rectified Flow and Our Rectified Diffusion.}
    \label{fig:overview}
\end{figure}

The motivation of this paper is to \textit{investigate what is most essential about rectified flow.} We argue that the effectiveness of rectified flow stems from using a pretrained diffusion model to acquire matched pairs of noise and samples, followed by retraining with these matched noise-sample pairs (i.e., the aforementioned third point). Based on this, the aforementioned first two points~(i.e., flow-matching \&  $\boldsymbol{v}$-prediction) are unnecessary. This allow us generalize the design space of rectified flow and make it adoptable for different diffusion variants including DDPM~\citep{ddpm}, EDM~\citep{edm}, Sub-VP~\citep{sde}, and etc. 

To this end, we propose Rectified Diffusion, as illustrated in Fig.~\ref{fig:overview}, our overall design is straightforward. We keep everything of the pretrained diffusion models unchanged, including noise schedulers, prediction types, networks architectures, and even training and inference code. The only difference is that the noise $\boldsymbol \epsilon$ and data $\mathbf x_0$ adopted for training are pre-collected and generated by the pretrained diffusion models instead of independently sampled from Gaussian and real data datasets.  

Additionally, \textit{we highlight that straightness is no more an essential training target} when we generalize the design space from solely flow-matching to more general diffusion forms. We analyze and show that the training target  of Rectified Diffusion is to obtain a \textit{first-order approximate ODE path}.  In simple terms, a first-order ODE implies the predictions of models remain consistent along the ODE trajectory and it still maintains at the same ODE trajectory after each denoising step. For models like DDPM~\citep{ddpm}, the first-order ODE path is inherently curved instead of straight. Therefore, `straightness' is no more suitable for Rectified Diffusion and is just a special case when we use the form of flow-matching.  

\begin{figure}
    \centering
    \vspace{-0.2in}
\includegraphics[width=0.8\linewidth]{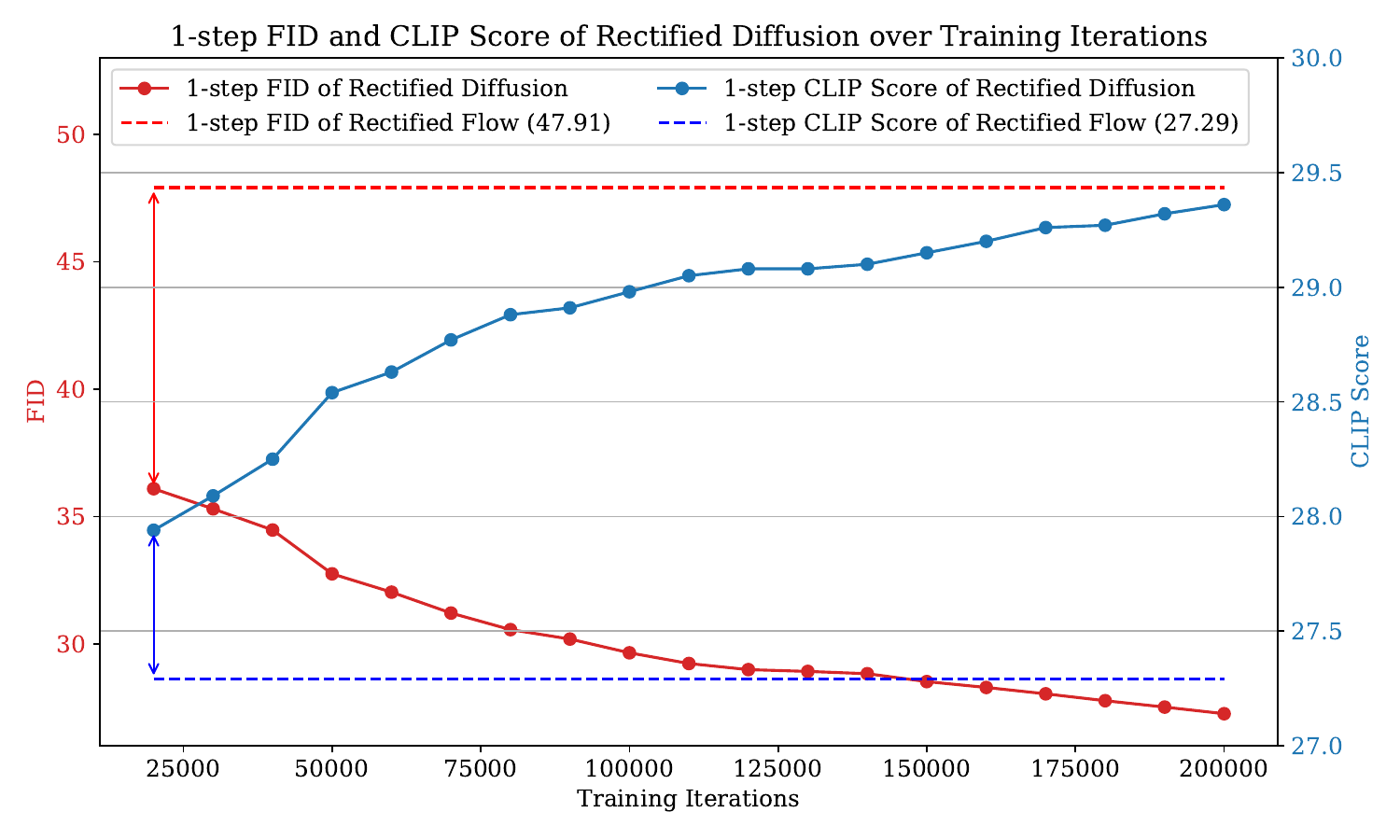}
    \vspace{-1em}
    \caption{Training iterations. 1-step performance of Rectified Diffusion significantly surpasses the 1-step performance of Rectified Flow within only 20,000 iterations with batch size 128~(8\% trained images of Rectified Flow) and consistently grows with more training iterations.}
    \label{fig:training-iterations}
    \vspace{-0.1in}
\end{figure}

To empirically validate our claim, we conduct experiments using Stable Diffusion, comparing our approach with InstaFlow~\citep{instaflow}, a key baseline based on rectified flow for text-to-image generation. We adhere the training setting of InstaFlow. The primary distinction is that InstaFlow requires transforming the Stable Diffusion models into a $\boldsymbol v$-prediction flow-matching model, while our method leaves everything of original Stable Diffusion unchanged. Our results demonstrate apparently better performance and faster training, likely due to our minimal differences in diffusion configurations. Our one-step performance achieves significantly superior performance  with only 8\% trained images of InstaFlow as shown in Fig.~\ref{fig:training-iterations}.  

Besides, we propose to replace the second-stage distillation adopted in InstaFlow with consistency distillation. We observe that the first-order approximate ODE path greatly facilitates consistency distillation, allowing us to achieve better performance at 3\% the GPU days than the further distilled model of InstaFlow. Additionally, we introduce Rectified Diffusion~(Phased), which divides the ODE path along the time axis into multiple segments and enforces first-order linearity within each segment. While this segmentation increases the minimum number of generation steps to match the number of segments, it substantially reduces both training cost and time. When compared to the previous segment-based rectified flow method, PeRFlow~\citep{yan2024perflow}, our approach demonstrates significantly better performance in experiments conducted on Stable Diffusion v1-5~\citep{rombach2022high} and Stable Diffusion XL~\citep{sdxl}.

We summarize our main contributions as follows: \textbf{(i)} We conduct an in-depth analysis of the essence of rectification and extend rectified flow to rectified diffusion. \textbf{(ii)} We identify that it is not straightness but first-order property is the essential training target of rectified diffusion with theoretical derivations.  \textbf{(iii)} Comprehensive comparisons on rectification, distillation and phased OED segmentation demonstrate our method achieves superior trade-off between generation quality and training efficiency over rectified flow-based models.

%% file: secs/sec3_analysis.tex
\section{Rectified Diffusion: Generalizing the design space of rectified flow into general diffusion models}

\paragraph{Rectified flow is a subset of rectified diffusion.} In the following discussion, we apply the diffusion form $\mathbf x_t = \alpha_t \mathbf x_0 + \sigma_t \boldsymbol \epsilon$ to introduce rectified diffusion. Note that this form of diffusion covers the flow-matching since we can simply set $\alpha_t =1-t$ and $\sigma_t = t$. Considering the different prediction types, we apply the epsilon-prediction for the following discussion. But note that different prediction types can be converted effortlessly through re-parameterization. For $\mathbf x_0$-prediction, we have $\mathbf x_0 = \frac{\mathbf x_t - \sigma_t\boldsymbol \epsilon}{\alpha_t}$. For $\boldsymbol v$-prediction utilized in rectified flow, we have $\boldsymbol v = \mathbf x_0 - \boldsymbol \epsilon = \frac{\mathbf x_t - (\alpha_t + \sigma_t) \boldsymbol \epsilon}{\alpha_t}$.  
Hence, we claim that \textit{rectified flow is a subset of rectified diffusion, and rectified diffusion is a generalization of rectified flow.}

\subsection{The nature of rectification is the retraining with pre-computed noise-sample pair}  \paragraph{The secret of rectification is using paired noise-sample for training.} To illustrate the differences clearly, we visualize the training processes for standard flow matching and rectification (reflow) training, as described in Algorithm~\ref{alg:alg1} and Algorithm~\ref{alg:alg3}, respectively. Differences are highlighted in red. A key observation is that in standard flow matching training, \(\mathbf{x}_0\) represents real data randomly sampled from the training set, while the noise \(\boldsymbol{\epsilon}\) is also randomly sampled from Gaussian. This results in random pairing between noise and sample. In contrast, in rectification training, the noise is pre-sampled from Gaussian, and the images are generated using pre-sampled noise by the model from the previous round of rectification (the pre-trained model), leading to a deterministic pairing.

\paragraph{Flow-matching training is a subset of standard diffusion training.} In addition, Algorithm~\ref{alg:alg2} visualizes the training process of a more general diffusion model, with differences to Algorithm~\ref{alg:alg1} highlighted in blue and orange. It’s important to note that flow matching is a specific case of the diffusion forms we discuss. From the algorithms, it is evident that the only distinctions between them lie in the diffusion form and prediction type. Consequently, flow matching training is just a special case of general diffusion training under a particular diffusion form and prediction type.

By comparing Algorithms~\ref{alg:alg2} and \ref{alg:alg3} with Algorithm~\ref{alg:alg1}, it is straightforward to derive Algorithm~\ref{alg:alg4}. Essentially, by incorporating the pre-trained model to collect noise-sample pairs and replacing the randomly sampled noise and real samples with these pre-collected pairs in the general diffusion training, we obtain the training algorithm for rectified diffusion.

\begin{multicols}{2}

\begin{algorithm}[H]
\caption{Flow Matching $\boldsymbol v$-Prediction}\label{alg:alg1}
\begin{algorithmic}
\State \textbf{Input:} 

\State Sample $\mathbf{x}_0$ from the data distribution
\State Sample time $t$ from a predefined schedule or uniformly from $[0, 1]$
\State Sample noise $\boldsymbol{\epsilon}$ from normal distribution
\State \textcolor{orange}{Compute $\mathbf{x}_t$ : $\mathbf{x}_t = (1 - t) \cdot \mathbf{x}_0 + t \cdot \boldsymbol{\epsilon}$}
\State \textcolor{orange}{Predict velocity $\hat{\boldsymbol{v}}$ using the model: $\hat{\boldsymbol{v}} = \text{Model}(\mathbf{x}_t, t)$}
\State \textcolor{orange}{Compute loss: $\mathcal L = \| \hat{\boldsymbol{v}} - (\mathbf{x}_0 - \boldsymbol \epsilon )\|_2^2$}
\State Backpropagate and update parameters
\end{algorithmic}
\end{algorithm}

\begin{algorithm}[H]
\caption{Diffusion Training $\boldsymbol \epsilon$-Prediction}\label{alg:alg2}
\begin{algorithmic}
\State \textbf{Input:}   \textcolor{blue}{$\alpha_t$, $\sigma_t$}

\State Sample $\mathbf{x}_0$ from the data distribution
\State Sample time $t$ from a predefined schedule or uniformly from $[0, 1]$
\State Sample noise $\boldsymbol{\epsilon}$ from normal distribution
\State \textcolor{blue}{Compute $\mathbf{x}_t$ : $\mathbf{x}_t = \alpha_t \cdot \mathbf{x}_0 + \sigma_t \cdot \boldsymbol{\epsilon}$}
\State \textcolor{blue}{Predict noise $\hat{\boldsymbol{\epsilon}}$ using the model: $\hat{\boldsymbol{\epsilon}} = \text{Model}(\mathbf{x}_t, t)$ }
\State \textcolor{blue}{Compute loss: $\mathcal L = \| \hat{\boldsymbol{\epsilon}} - \boldsymbol{\epsilon} \|_2^2$}
\State Backpropagate and update parameters
\end{algorithmic}
\end{algorithm}

\end{multicols}
\begin{multicols}{2}

\begin{algorithm}[H]
\caption{Rectified Flow $\boldsymbol v$-Prediction}\label{alg:alg3}
\begin{algorithmic}
\State \textbf{Input:} \textcolor{red}{noise-data pair $(\boldsymbol{\epsilon},\hat{\mathbf{x}}_0)$}

\State \sout{Sample $\mathbf{x}_0$ from the data distribution}
\State Sample time $t$ from a predefined schedule or uniformly from $[0, 1]$
\State \sout{Sample noise $\boldsymbol{\epsilon}$ from normal distribution}
\State \textcolor{orange}{Compute $\mathbf{x}_t$ :} $\textcolor{orange}{\mathbf{x}_t = (1 - t) \cdot }\textcolor{red}{\hat{\mathbf{x}}_0} \textcolor{orange}{ + t \cdot \boldsymbol{\epsilon}}$
\State \textcolor{orange}{Predict velocity $\hat{\boldsymbol{v}}$ using the model: $\hat{\boldsymbol{v}} = \text{Model}(\mathbf{x}_t, t)$}
\State \textcolor{orange}{Compute loss:} $\textcolor{orange}{\mathcal L = \| \hat{\boldsymbol{v}} -} (\textcolor{red}{\hat{\mathbf{x}}_0} \textcolor{orange}{- \boldsymbol \epsilon )\|_2^2}$
\State Backpropagate and update parameters
\end{algorithmic}
\end{algorithm}

\begin{algorithm}[H]
\caption{Rectified Diffusion $\boldsymbol \epsilon$-Prediction}\label{alg:alg4}
\begin{algorithmic}
\State \textbf{Input:} \textcolor{red}{noise-data pair $(\boldsymbol{\epsilon},\hat{\mathbf{x}}_0)$},  \textcolor{blue}{$\alpha_t$, $\sigma_t$}

\State \sout{Sample $\mathbf{x}_0$ from the data distribution}
\State Sample time $t$ from a predefined schedule or uniformly from $[0, 1]$
\State \sout{Sample noise $\boldsymbol{\epsilon}$ from normal distribution}
\State \textcolor{blue}{Compute $\mathbf{x}_t$ :} $\textcolor{blue}{\mathbf{x}_t = \alpha_t \cdot} \textcolor{red}{\hat{\mathbf{x}}_0} \textcolor{blue}{+ \sigma_t \cdot \boldsymbol{\epsilon}}$
\State \textcolor{blue}{Predict noise $\hat{\boldsymbol{\epsilon}}$ using the model: $\hat{\boldsymbol{\epsilon}} = \text{Model}(\mathbf{x}_t, t)$}
\State \textcolor{blue}{Compute loss: $\mathcal L = \| \hat{\boldsymbol{\epsilon}} - \boldsymbol{\epsilon} \|_2^2$}
\State Backpropagate and update parameters
\end{algorithmic}
\end{algorithm}

\end{multicols}



\subsection{Understanding the first-order ODE~($\star\star\star$)}
For the above discussed general diffusion form $\mathbf x_t = \alpha_t \mathbf x_0 + \sigma_t \boldsymbol \epsilon$, there exists an exact ODE solution form~\citep{dpmsolver},
\begin{align}\label{eq:exact}
    \mathbf x_t = \frac{\alpha_t}{\alpha_s} \mathbf x_s - \alpha_t \int_{\lambda_s}^{\lambda_t} e^{-\lambda}\textcolor{blue}{ \boldsymbol \epsilon_{\boldsymbol \theta} (\mathbf x_{t_{\lambda}}, {t_{\lambda}})} \mathrm d \lambda\, ,
\end{align}
where $\lambda_t = \ln \frac{\alpha_t}{\sigma_t}$, and $t_{\lambda}$ is the inverse function of $\lambda_t$. The left term $\frac{\alpha_t}{\alpha_s} \mathbf x_s$ is a pre-defined deterministic scaling. The right term is the exponentially weighted integral of epsilon predictions. The first order ODE means the above integral with arbitrary $t$ and $s$ is equivalent to 
\begin{align}\label{eq:first-order}
        \mathbf x_t = \frac{\alpha_t}{\alpha_s} \mathbf x_s - \alpha_t \textcolor{red}{\boldsymbol \epsilon_{\boldsymbol \theta} (\mathbf x_{s}, s)}\int_{\lambda_s}^{\lambda_t} e^{-\lambda}  \mathrm d \lambda = \frac{\alpha_t}{\alpha_s} \mathbf x_s + \alpha_t \textcolor{red}{\boldsymbol \epsilon_{\boldsymbol \theta} (\mathbf x_{s}, s)} (\frac{\sigma_t}{\alpha_t} - \frac{\sigma_s}{\alpha_s})\, .
\end{align}
We show that the equivalent of Equation~\ref{eq:exact} and Equation~\ref{eq:first-order} for arbitrary $t$ and $s$ \textit{holds and only holds} if the epsilon prediction on the same ODE trajectory is a constant in Thereom~\ref{th:first-order}. 

\paragraph{First-order ODE has the same form of predefined diffusion form.} To put it in a nutshell, we assume the ODE trajectory is a first-order ODE, and there exists a solution point $\mathbf x_0$. Therefore, the epsilon predictions on the ODE trajectory with solution point $\mathbf x_0$ are constant, which we denote as $\boldsymbol \epsilon$. Substitute $s=0$, $\mathbf x_0$, $\alpha_s = 1$, $\sigma_s = 0$ and $\boldsymbol \epsilon$ into Equation~\ref{eq:first-order}, we have 
\begin{align}~\label{eq:first-order-gt}
    \mathbf x_t = \alpha_t \mathbf x_0 + \sigma_t \boldsymbol \epsilon. 
\end{align}
This has exactly the same form of predefined forward process.  Therefore, we have that the first-order ODE is exactly the weighted interpolation of data and noise following predefined forward diffusion form.  The only difference is that, the $\boldsymbol \epsilon$ and $\mathbf x_0$ in the above equation is deterministic pair on the same ODE trajectory. While, for standard diffusion training, the $\mathbf x_0$ and $\boldsymbol \epsilon$ are randomly sampled.
\textit{That indicates that if we achieve perfect coupling of data~$\mathbf x_0$ and noise~$\boldsymbol \epsilon$ at training, and there's no intersections among different paths~(otherwise the epsilon predictions can be the epsilon prediction expectation of different paths), the trained diffusion models in the ideal case~(without optimization error) will obtain the first-order ODE.} 

\paragraph{First-order ODE supports consistent generation with arbitrary inference steps.} Additionally, note that if the epsilon predictions on the same trajectory are constant, it is easy to show that the $\mathbf x_0$-predictions are also constant. Therefore, the first-order ODE can flexibly support one-step generation~($\mathbf x_T \rightarrow \mathbf x_0$) or multi-step generation~($\mathbf x_T \rightarrow \dots \rightarrow \mathbf x_0$).  If a perfect first-order ODE is achieved, we will always get indentical generation results with arbitrary inference steps.

\begin{figure}[t]
    \centering
    \begin{subfigure}[b]{0.31\textwidth}
        \centering
        \includegraphics[width=\textwidth]{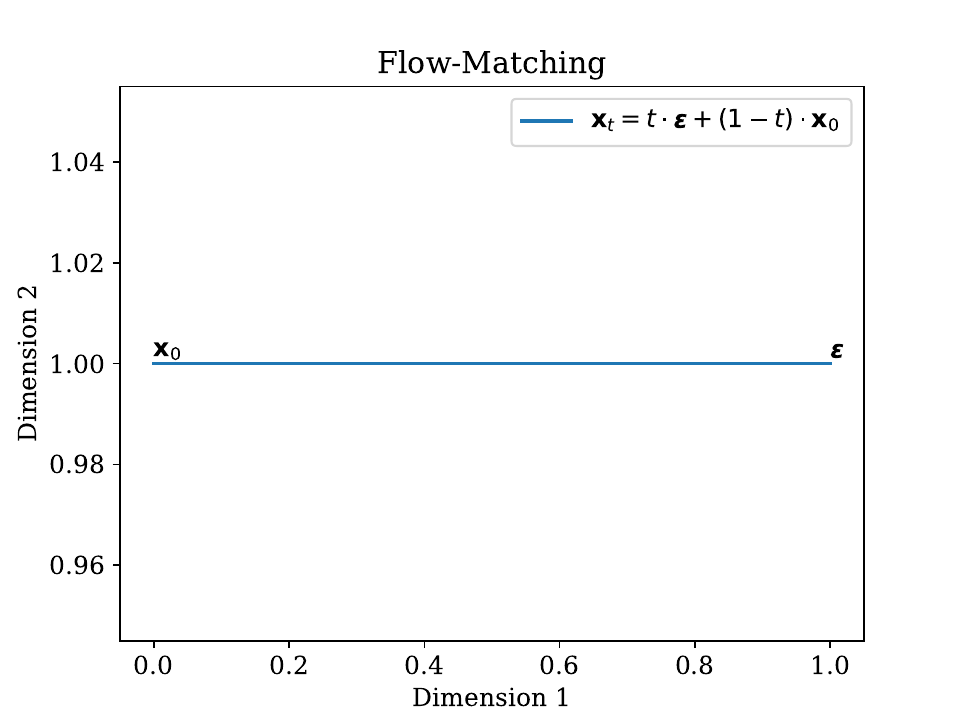}
        \caption{Flow Matching}
        \label{fig:fm}
    \end{subfigure}
    \hfill
    \begin{subfigure}[b]{0.31\textwidth}
        \centering
        \includegraphics[width=\textwidth]{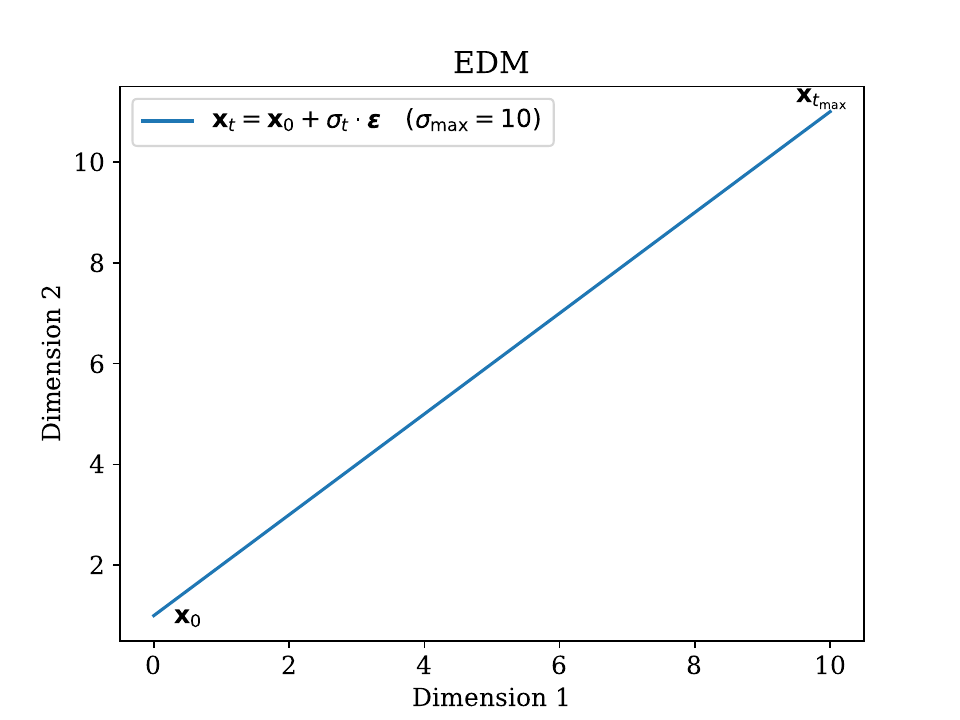}
        \caption{EDM}
        \label{fig:edm}
    \end{subfigure}
    \hfill
    \begin{subfigure}[b]{0.31\textwidth}
        \centering
        \includegraphics[width=\textwidth]{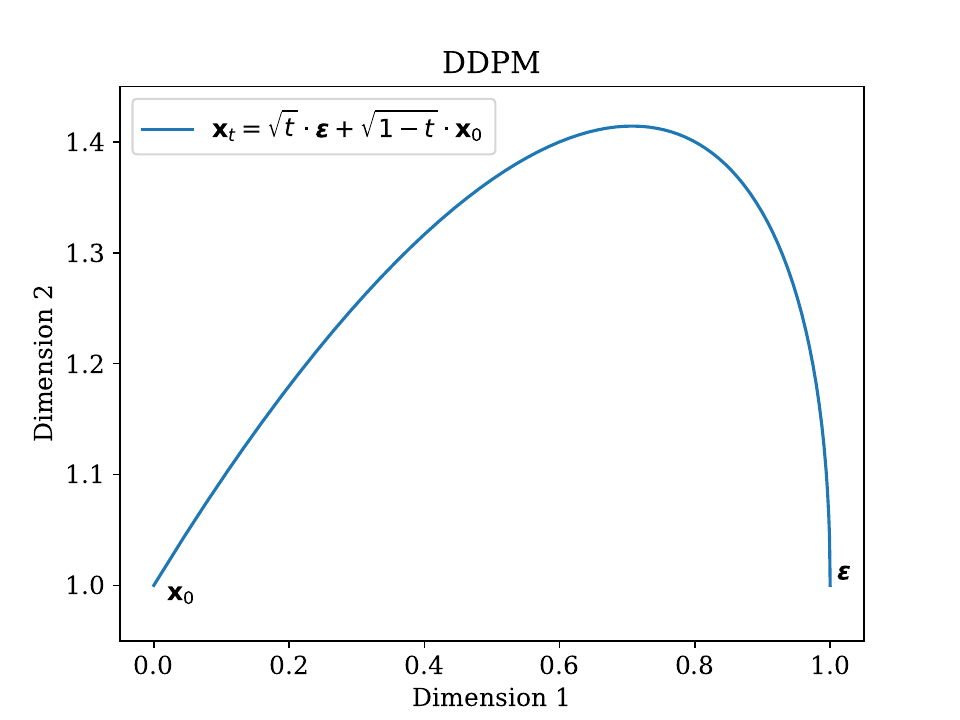}
        \caption{DDPM}
        \label{fig:ddpm}
    \end{subfigure}
    \hfill
    \begin{subfigure}[b]{0.31\textwidth}
        \centering
        \includegraphics[width=\textwidth]{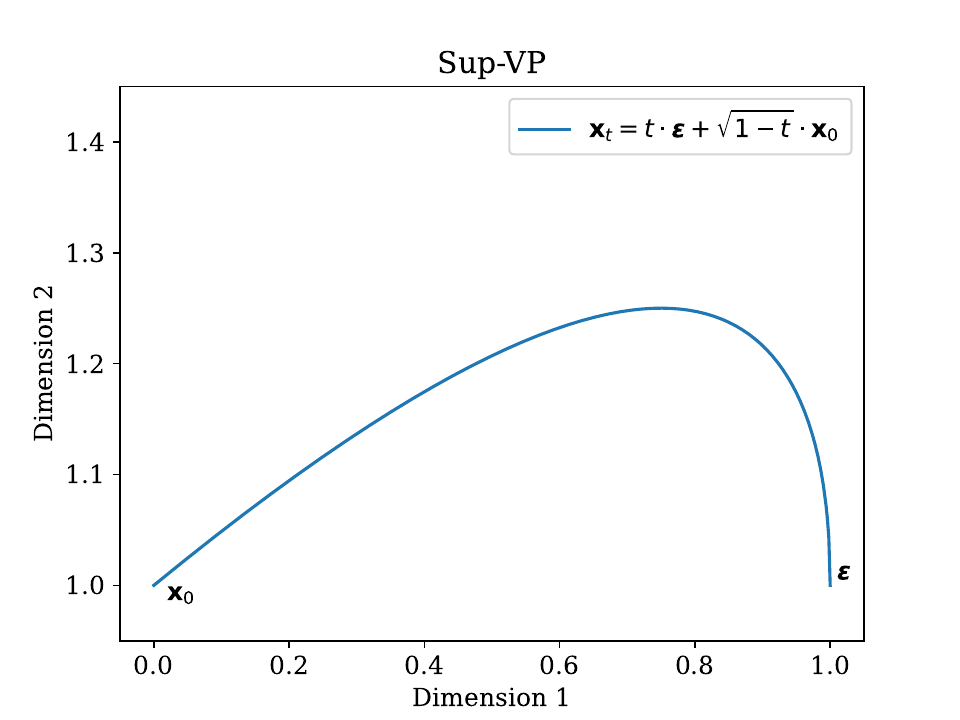}
        \caption{Sub-VP}
        \label{fig:subvp}
    \end{subfigure}
    \begin{subfigure}[b]{0.31\textwidth}
        \centering
        \includegraphics[width=\textwidth]{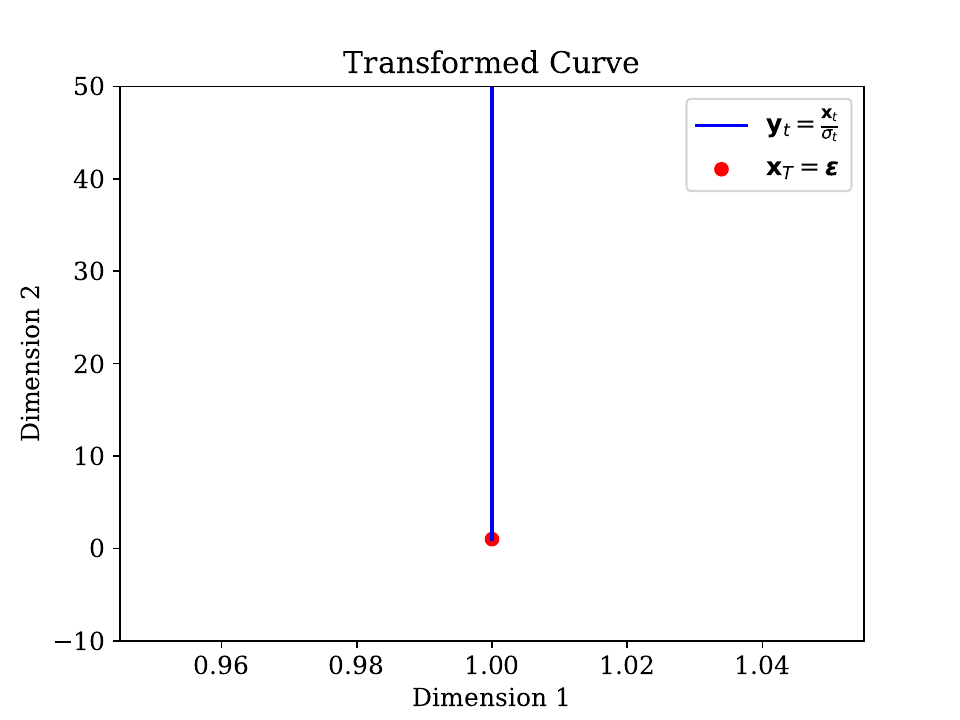}
        \caption{Transformed}
        \label{fig:transformed}
    \end{subfigure}
    \caption{First-order trajectory of different diffusion forms. We show that the first-order ODE has the same form as their predefined forward process, i.e., $\mathbf x_t = \alpha_t \mathbf x_0 + \sigma_t \boldsymbol \epsilon$. Though the first-order ODE paths of Flow Matching and EDM are straight, the first-order ODE paths of DDPM and Sub-VP are inherently curved. First-order ODE paths of all diffusion forms can be converted into straight lines through simple scaling as shown in Fig.~\ref{fig:transformed}.}
    \label{fig:forms}
\end{figure}

\paragraph{First-order ODE can be inherently curved.} For the first-order ODE, though the trajectories of flow-matching based methods are straight, the trajectories of other forms of diffusion models can be  inherently curved. But if we define $\mathbf y_t =\frac{\mathbf x_t}{\sigma_t}$, we will have $\mathbf y_t = \frac{\alpha_t}{\sigma_t}\mathbf x_0 + \boldsymbol \epsilon $ from the Equation~\ref{eq:first-order-gt}. We can easily obeserve that the trajectory of $\mathbf y_t$ is a straight line from the initial point $\boldsymbol \epsilon$ towards the direction of $\mathbf x_0$~(i.e, first-order trajectories can be converted to straight lines). We showcase our findings in Fig.~\ref{fig:forms}.We select $\mathbf x_0 = [0, 1]$ and $\boldsymbol \epsilon = [1, 1]$ The Fig.~\ref{fig:fm} and Fig.~\ref{fig:edm} show the first-order trajectory of flow-matching and EDM. They are both straight, but EDM has a totally different trajectory and magnitude. Fig.~\ref{fig:ddpm} and Fig.~\ref{fig:subvp} show the first-order trajectory of DDPM and Sub-VP. Their first-order trajectory are inherently curved. Fig.~\ref{fig:transformed} shows the trajectory of $\mathbf y_t = \frac{\alpha_t}{\sigma_t}\mathbf x_0 + \boldsymbol \epsilon$. It shows that all the first-order trajectories can be converted into straight lines with simple timestep-dependent scaling.

\subsection{Rectified Diffusion~(Phased)}

\begin{figure}[t]
    \centering
    \includegraphics[width=0.8\linewidth]{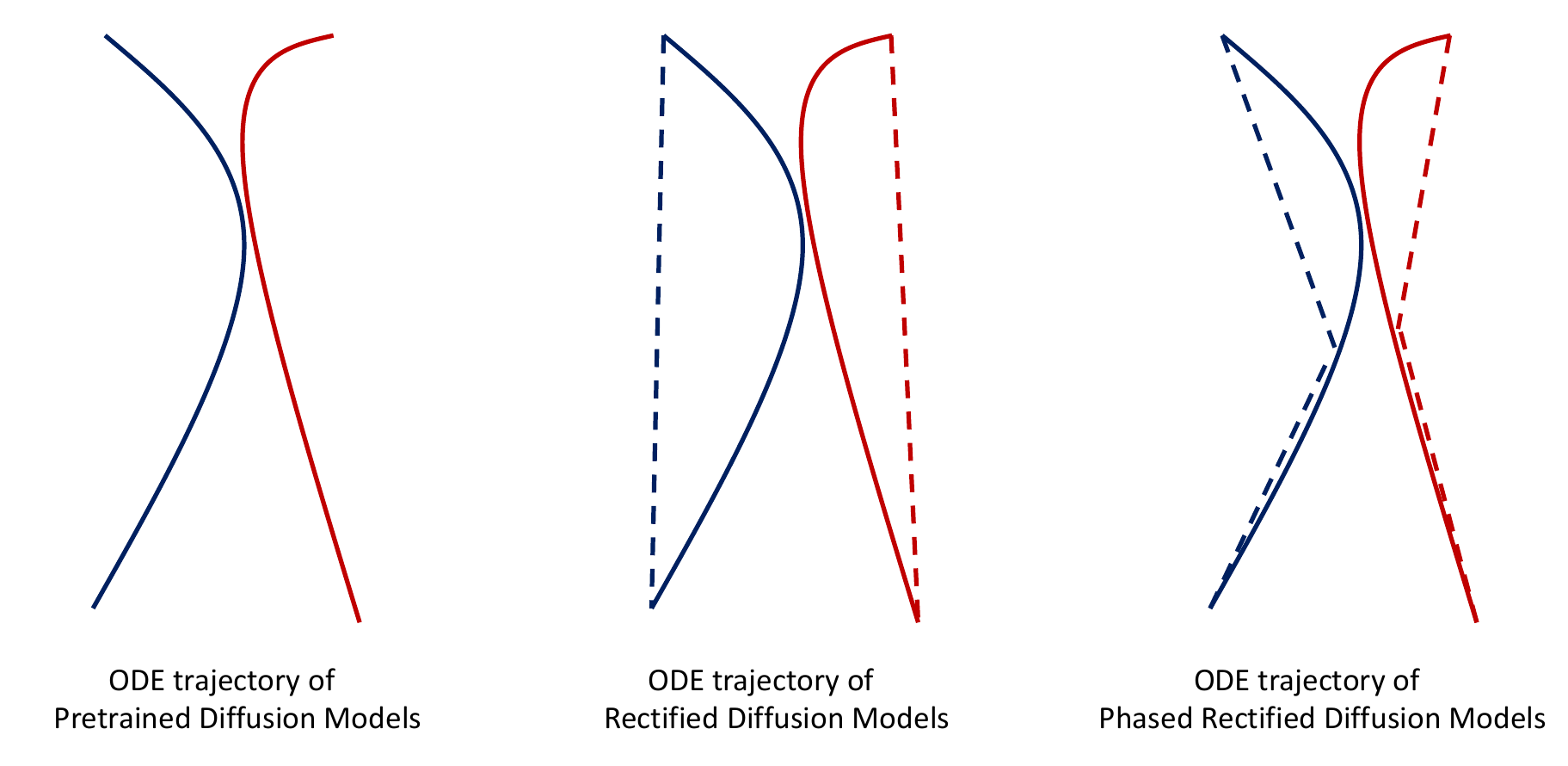}
    \caption{ODE trajectory comparison of diffusion models, rectified diffusion models, and phased consistency models. We apply straight lines for more clear demonstration. The solid line shows the original diffusion ODE path, while the dashed line shows the rectified ODE path.}
    \label{fig:compare}
\end{figure}

Completely linearizing the ODE path of a pre-trained diffusion model is challenging because the original ODE can deviate significantly from a first-order form. In Fig.~\ref{fig:compare}, we visualize both the original diffusion ODE path and the corresponding first-order ODE path. Since it's hard to intuitively determine whether a curved ODE path satisfies first-order linearity, we represent the first-order ODE path with a straight line. A significant gap between the two paths is evident. However, enforcing local first-order linearity is more feasible. As shown on the right side of the figure, when the ODE path is divided into two segments along the time axis and each segment is linearized separately, the new ODE path is closer to the original one. This observation motivates the development of our rectification diffusion~(phased).

We set intermediate time steps as $s_0 = 0 < s_1 < s_2 < \cdots < s_{M-1} = t_{\max}$ along the time axis of ODE, where $M$ is the number of phases. The training process begins with sampling $\mathbf x_0$ from real data, followed by adding random noise at time step $s_m$ to obtain $\mathbf x_{s_m}$. We then use the pre-trained diffusion model to perform multi-step numerical solving  to obtain $\mathbf x_{s_{m-1}}$ for the previous intermediate step. However, the phasing idea involves two challenges: 1) determining the noise $\boldsymbol \epsilon$ corresponding to the first-order path, and 2) determine the sample $\mathbf x_t$ at any time $t$ between $s_m$ and $s_{m-1}$ on the same first-order ODE, where $t \in (s_{m-1}, s_m)$.

Fortunately, the transition formula between any two points on the first-order ODE is known, as shown in Equation~\ref{eq:first-order}. Through a simple transformation, we have the noise $\boldsymbol \epsilon$ corresponding to the ODE path between $\mathbf x_{s_m}$ and $\mathbf x_{s_{m-1}}$ can be expressed as:
\begin{align}
\boldsymbol \epsilon = \frac{\frac{\mathbf x_{s_{m-1}}}{\alpha_{s_{m-1}}} - \frac{\mathbf x_{s_{m}}}{\alpha_{s_{m}}}}{\frac{\sigma_{s_{m-1}}}{\alpha_{s_{m-1}}} -\frac{\sigma_{s_{m}}}{\alpha_{s_{m}}}} = \frac{\Delta \mathbf z}{\Delta \text{NSR}}\, ,
\end{align}
where $\Delta \mathbf z$ represents the change in $\mathbf z_t = \frac{\mathbf x_t}{\alpha_t}$, and $\Delta \text{NSR}$ denotes the change in $\frac{\sigma_t}{\alpha_t}$.
Once this noise $\boldsymbol \epsilon$ is calculated, it can be directly substituted into Equation~\ref{eq:first-order} to compute $\mathbf x_t$ at any time $t$ along the ODE path.

\subsection{Rectified Diffusion facilitates the consistency distillation}
\begin{figure}[t]
    \centering
    \begin{subfigure}[b]{0.31\textwidth}
        \centering
        \includegraphics[width=\textwidth]{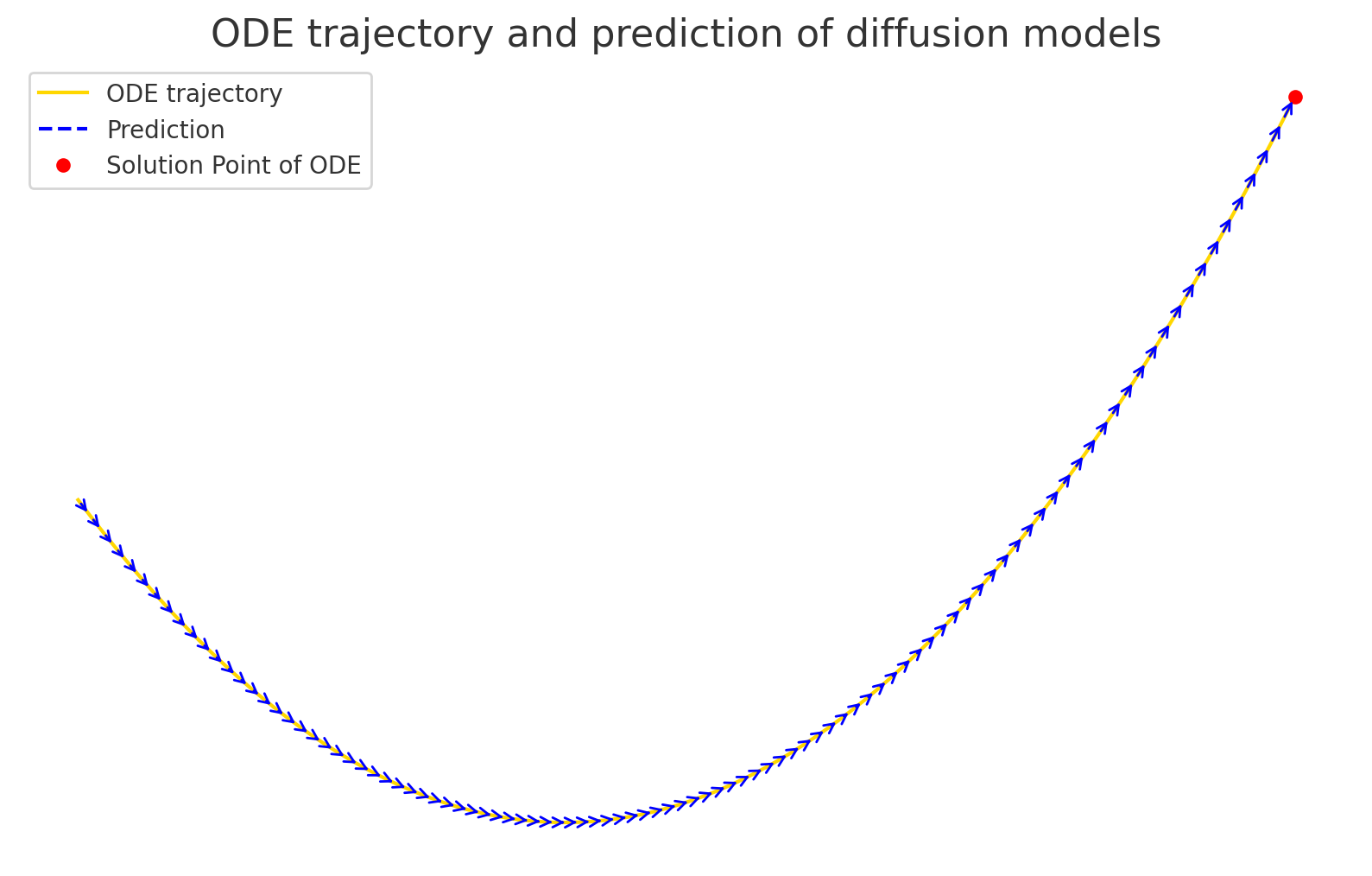}
        \caption{Diffusion models}
        \label{fig:dm}
    \end{subfigure}
    \hfill
    \begin{subfigure}[b]{0.31\textwidth}
        \centering
        \includegraphics[width=\textwidth]{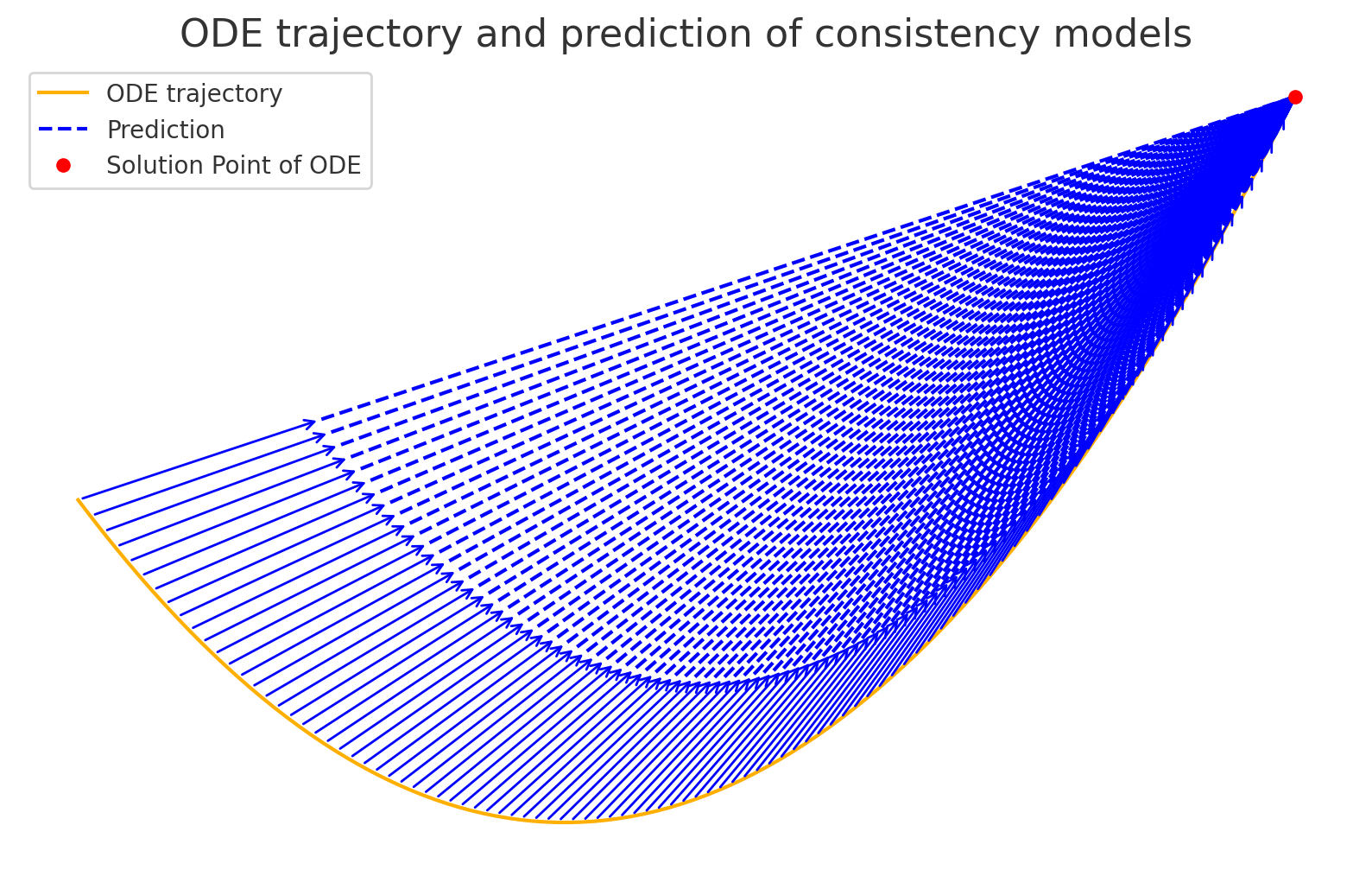}
        \caption{Consistency models}
        \label{fig:cm}
    \end{subfigure}
    \hfill
    \begin{subfigure}[b]{0.31\textwidth}
        \centering
        \includegraphics[width=\textwidth]{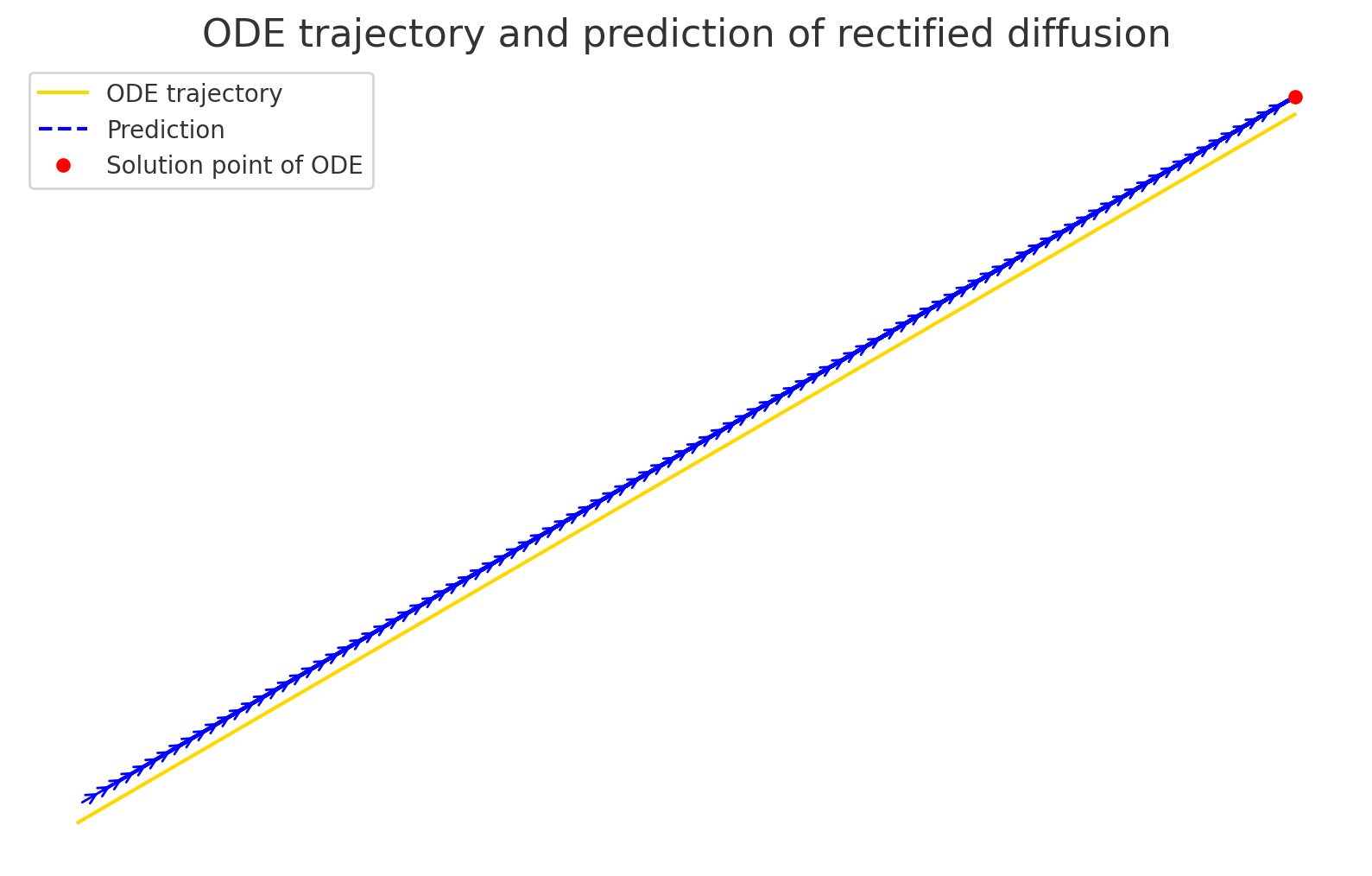}
        \caption{Rectified diffusion~(Ours)}
        \label{fig:rectified-diffusion}
    \end{subfigure}
    \caption{ODE trajectory and prediction comparison of consistency models and reftified diffusion. We apply straight lines for more clear demonstration. The yellow line shows the ODE trajectories, while the blue line shows the predictions.}
    \label{fig:cm-rectified-diffusion}
\end{figure}

Previous work~\citep{instaflow} proposes applying naive distillation after rectification to enhance one-step generation ability. This is because, after rectification, the model cannot achieve a perfect first-order path due to issues like optimization, model capacity, and ODE path intersections. As a result, rectified flow-based methods still do not perform as well as the most advanced distillation methods at low-step regime~(e.g., 1-step generation). Following it, we also utilize distillation to further improve the model's performance at low-step regime after rectified diffusion. Instead of using naive distillation, we employ the more advanced distillation technique--consistency distillation~\citep{cm}, which eliminates the need to regenerate large numbers of samples. Moreover, we found that after rectification, where the ODE path is approximately first-order, consistency distillation leads to significantly faster training and better performance. This is because the training objective of a first-order ODE imposes a stronger constraint than self-consistency. In Fig.~\ref{fig:cm-rectified-diffusion}, we illustrate the differences between the diffusion model, consistency model, and rectified diffusion. The consistency model only adjusts the direction of the model's predictions without altering the ODE path itself, while rectified diffusion enforces a change in the ODE path to a first-order form.

%% file: secs/sec5_exp.tex
\section{Empirical Validation}

\begin{figure}
    \centering
\includegraphics[width=0.95\linewidth]{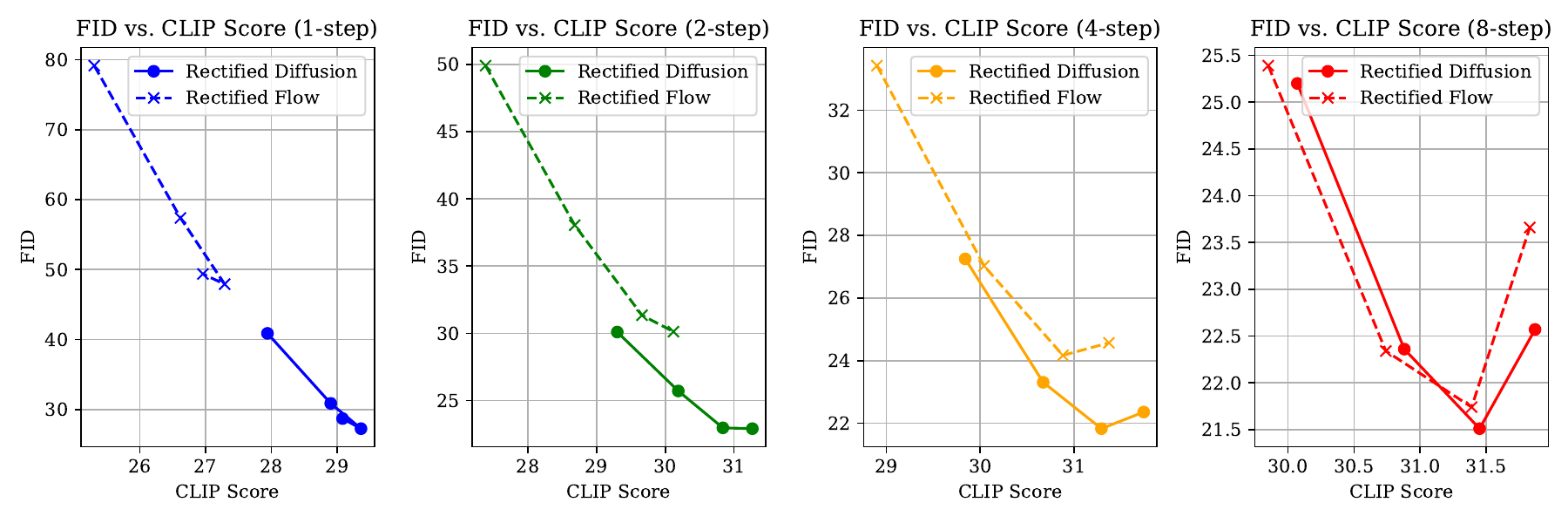}
    \caption{Effectiveness of Classifier-Free Guidance.}
    \label{fig:cfg}
\end{figure}

\subsection{Validation Setup}

To thoroughly compare our approach with rectified flow-based methods, we organize the comparison into three levels:

\begin{enumerate}[\arabic{enumi})]
    \item \textbf{Rectification Comparison}: InstaFlow~\citep{instaflow} proposes initializing a v-prediction-based flow-matching model using Stable Diffusion v1-5~\citep{rombach2022high}, followed by further training with their rectified flow method, which we refer to as Rectified Flow. To compare with this, we apply the rectified diffusion method to continue training Stable Diffusion v1-5, referred to as Rectified Diffusion. This comparison aims to demonstrate the faster training speed and superior performance of our proposed rectified diffusion approach.

    \item \textbf{Distillation Comparison}: In the InstaFlow paper, the authors suggest using a standard distillation technique to improve the model's performance in a one-step scenario, which we refer to as Rectified Flow (Distill). Similarly, we apply a distillation strategy to enhance performance at low-step regimes, specifically using consistency distillation to  boost training efficiency. This approach is termed Rectified Diffusion (CD).

    \item \textbf{Phased ODE Segmentation}: PeRFlow~\citep{yan2024perflow} introduces the concept of segmenting the ODE and presents experimental results on both SD and SDXL~\citep{sdxl}, termed PeRFlow and PeRFlow-XL, respectively. We extend this idea by proposing a method for phasing the ODE to enforce first-order property within each sub-phase, which we call Rectified Diffusion (Phased) and Rectified Diffusion-XL (Phased).
\end{enumerate}

Across all three of these comparative experiments, our methods demonstrate significantly superior performance.

\begin{table}[h]
\centering
\small
\caption{\small Performance comparison on validation set of COCO-2017.}
\label{tab:coco-2017}
\resizebox{.95\textwidth}{!}{
\begin{tabular}{@{}llccccc@{}}
\toprule
\textbf{Method} & \textbf{Res.} & \textbf{Time ($\downarrow$)} & \textbf{\# Steps} & \textbf{\# Param.} & \textbf{FID ($\downarrow$)} & \textbf{CLIP ($\uparrow$)}\\ \midrule
SDv1-5+DPMSolver~(Upper-Bound)~\citep{dpmsolver} & 512& 0.88s & 25 & 0.9B & 20.1 & 0.318\\\midrule
Rectified Flow~\citep{instaflow} & 512& 0.88s & 25 & 0.9B & 21.65 & 0.315\\ 
Rectified Flow~\citep{instaflow} & 512& 0.09s & 1 & 0.9B & 47.91 & 0.272\\ 
Rectified Flow~\citep{instaflow} & 512& 0.13s & 2 & 0.9B & 31.35 & 0.296\\ 
Rectified Diffusion~(Ours) & 512& 0.09s & 25 & 0.9B & 21.28 & 0.316\\ 
Rectified Diffusion~(Ours) & 512& 0.09s & 1 & 0.9B & 27.26 & 0.295\\ 
Rectified Diffusion~(Ours) & 512& 0.13s & 2 & 0.9B & 22.98 & 0.309\\ \midrule
Rectified Flow~(Distill)~\citep{instaflow}& 512& 0.09s & 1 & 0.9B & 23.72 & 0.302\\ 
Rectified Flow~(Distill)~\citep{instaflow}& 512& 0.13s & 2 & 0.9B & 73.49 & 0.261\\ 
Rectified Flow~(Distill)~\citep{instaflow}& 512& 0.21s & 4 & 0.9B & 103.48 & 0.245\\ 
Rectified Diffusion~(CD)~(Ours)& 512& 0.09s & 1 & 0.9B & 22.83 & 0.305\\ 
Rectified Diffusion~(CD)~(Ours)& 512& 0.13s & 2 & 0.9B & 21.66 & 0.312\\
Rectified Diffusion~(CD)~(Ours)& 512& 0.21s & 4 & 0.9B & 21.43 & 0.314\\
\midrule
PeRFlow~\citep{yan2024perflow}& 512& 0.21s & 4 & 0.9B & 22.97 & 0.294\\ 
Rectified Diffusion~(Phased)~(Ours)& 512& 0.21s & 4 & 0.9B & 20.64 & 0.311\\ 
\midrule 
PeRFlow-SDXL~\citep{yan2024perflow}& 1024& 0.71s & 4 & 3B & 27.06 & 0.335\\ 
Rectified Diffusion-SDXL~(Phased)~(Ours)& 1024& 0.71s & 4 & 3B & 25.81 & 0.341\\ 
 \bottomrule
\end{tabular}}
\end{table}

\begin{figure}[ht]
    \centering
    \begin{subfigure}[b]{0.31\textwidth}
        \centering
        \includegraphics[width=\textwidth]{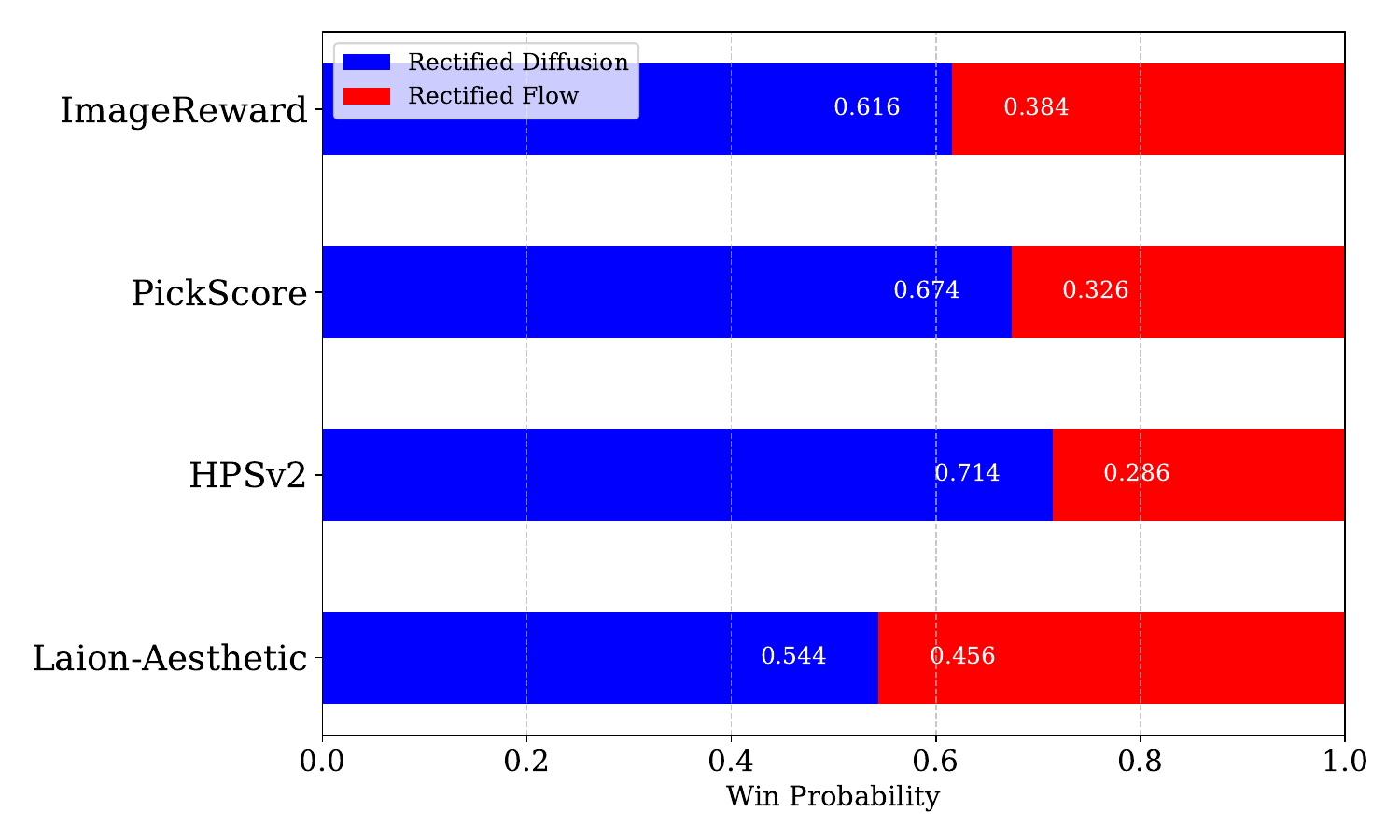}
        \caption{1-step comparison}
        \label{fig:hps-a}
    \end{subfigure}
    \hfill
    \begin{subfigure}[b]{0.31\textwidth}
        \centering
        \includegraphics[width=\textwidth]{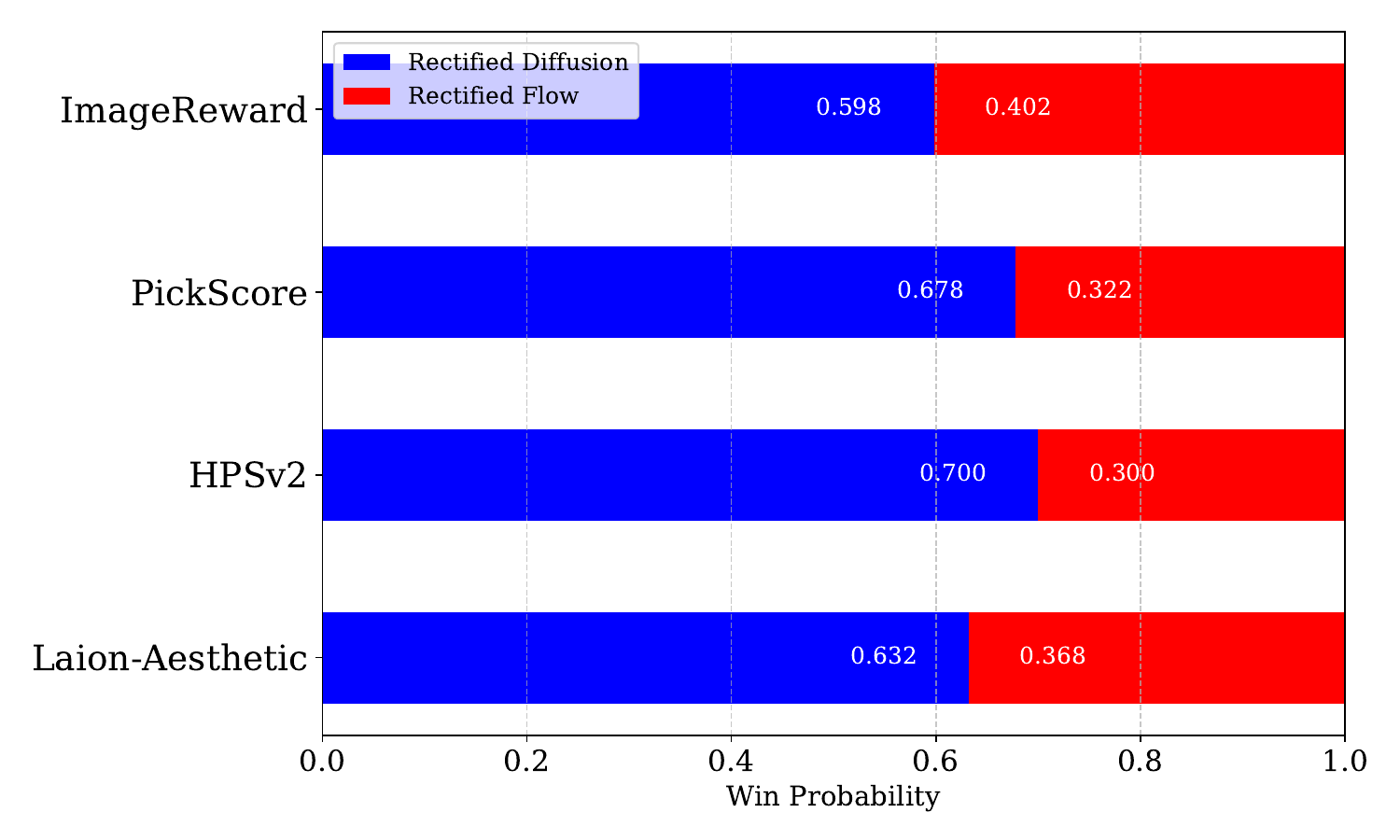}
        \caption{2-step comparison}
        \label{fig:hps-b}
    \end{subfigure}
    \hfill
    \begin{subfigure}[b]{0.31\textwidth}
        \centering
        \includegraphics[width=\textwidth]{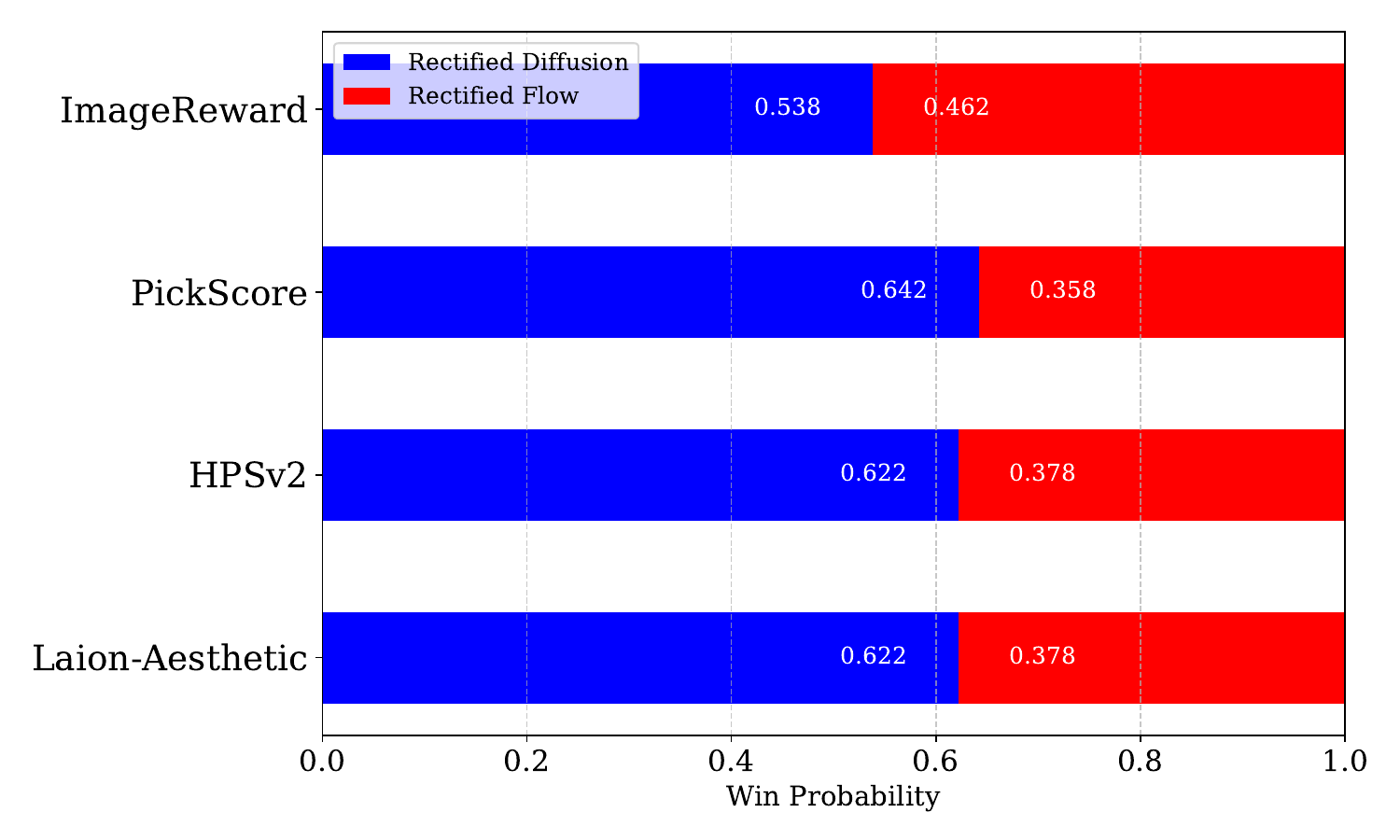}
        \caption{4-step comparison}
        \label{fig:hps-c}
    \end{subfigure}
    \hfill
    \begin{subfigure}[b]{0.31\textwidth}
        \centering
        \includegraphics[width=\textwidth]{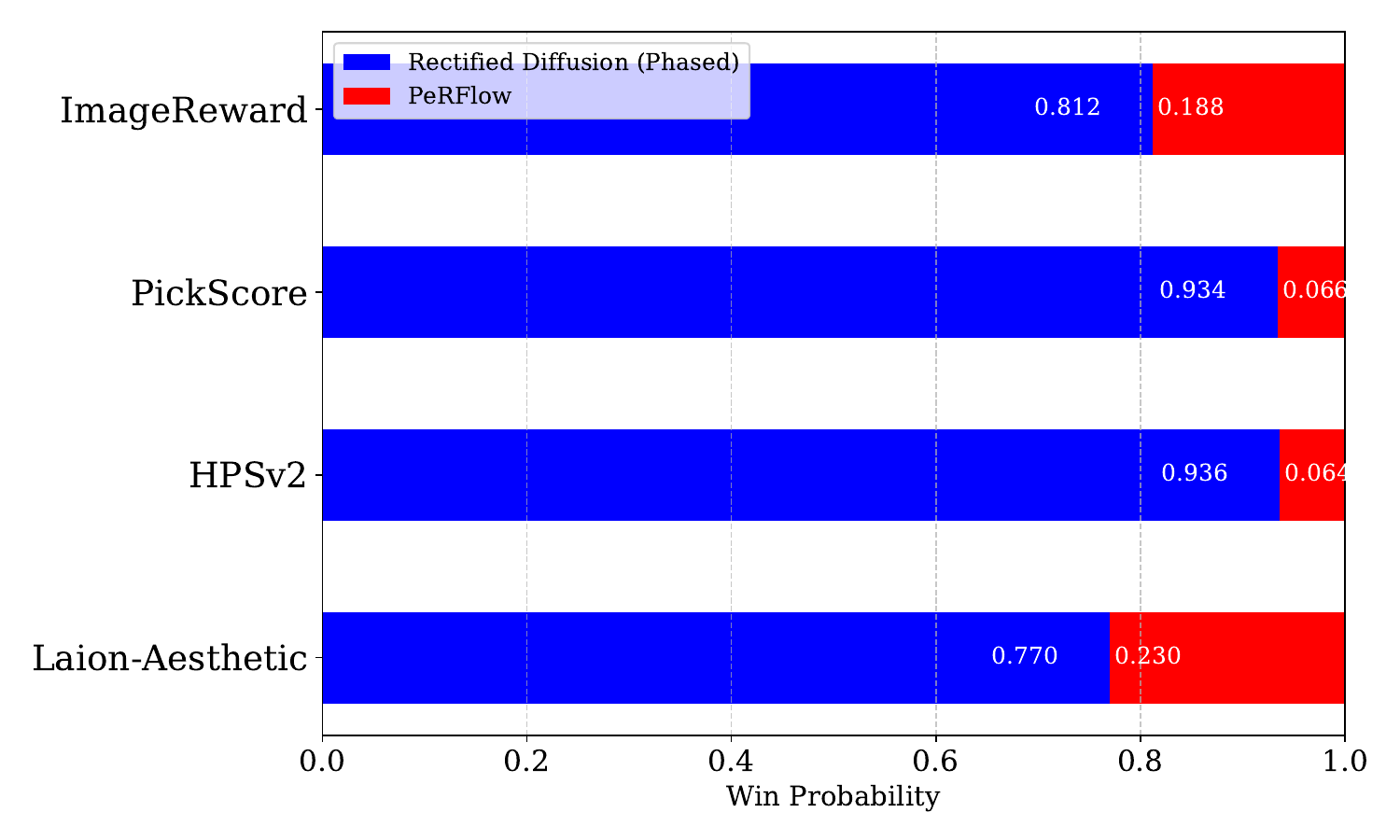}
        \caption{Phased~(4-step)}
        \label{fig:hps-d}
    \end{subfigure}
    \begin{subfigure}[b]{0.31\textwidth}
        \centering
        \includegraphics[width=\textwidth]{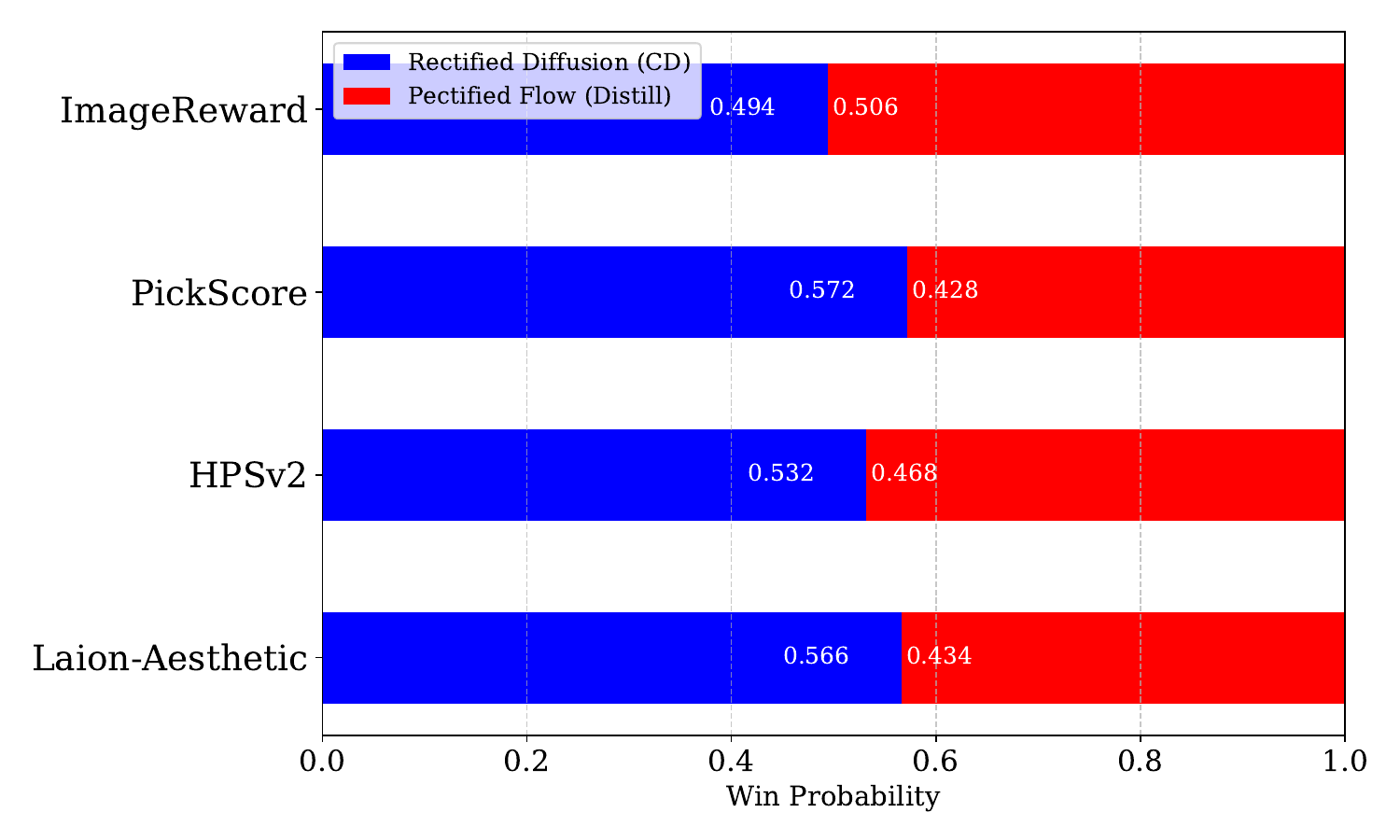}
        \caption{Distillation~(1-step)}
        \label{fig:hps-e}
    \end{subfigure}
    \caption{Human preference metrics comparison.}
    \label{fig:human}
\end{figure}

\begin{table}[htp]
\vspace{-0.3in}

\centering
\small
\caption{\small Performance comparison on COCO-2014.}
\label{tab:comparison}
\resizebox{.93\textwidth}{!}{
\begin{tabular}{@{}llccccc@{}}
\toprule
\textbf{Method} & \textbf{Res.} & \textbf{Time ($\downarrow$)} & \textbf{\# Steps} & \textbf{\# Param.} & \textbf{FID ($\downarrow$)} & \textbf{CLIP ($\uparrow$)}\\ \midrule
\multicolumn{7}{c}{Autoregressive Models} \\ 
 DALL·E~\citep{ramesh2021zero} & 256 & - & - & 12B & 27.5 & - \\
 CogView2~\citep{ding2021cogview} & 256 & - & - & 6B & 24.0 & - \\
 Parti-750M~\citep{yu2022scaling} & 256 & - & - & 750M & 10.71 & - \\
Parti-3B~\citep{yu2022scaling} & 256 & 6.4s & - & 3B & 8.10 & - \\
Parti-20B~\citep{yu2022scaling} & 256 & - & - & 20B & 7.23 & - \\
Make-A-Scene~\citep{gafni2022make} & 256 & 25.0s & - & - & 11.84 & - \\
\multicolumn{7}{c}{Masked Models} \\ 
Muse~\citep{chang2023muse}  & 256  &  1.3 & 24 & 3B & 7.88 & 0.32\\
\midrule
\multicolumn{7}{c}{Diffusion Models} \\ 
 GLIDE~\citep{nichol2021glide} & 256 & 15.0s & 250 & 5B & 12.24 & -\\
 DALL·E 2~\citep{ramesh2022hierarchical} & 256 & - & 250+27 & 5.5B & 10.39 & - \\
LDM~\citep{rombach2022high} & 256 & 3.7s & 250 & 1.45B & 12.63& - \\
Imagen~\citep{saharia2022photorealistic} & 256 & 9.1s & - & 3B & 7.27 & - \\
eDiff-I~\citep{balaji2022ediffi} & 256 & 32.0s & 25+10 & 9B & 6.95 & - \\
\midrule
\multicolumn{7}{c}{Generative Adversarial Networks (GANs)} \\
LAFITE~\citep{zhou2022towards} & 256 & 0.02s & 1 & 75M & 26.94 & - \\
 StyleGAN-T~\citep{sauer2023stylegan} & 512 & 0.10s & 1 & 1B & 13.90& $\sim$0.293\\
 GigaGAN~\citep{kang2023scaling} & 512 & 0.13s & 1 & 1B & 9.09 & -\\
  \midrule
  \multicolumn{7}{c}{Stable Diffusion~(0.9 B) and its accelerated or distilled versions}\\
  \multicolumn{7}{c}{GANs} \\ 
 UFOGen~\citep{xu2024ufogen} & 512 & 0.09s & 1 & 0.9B & 12.78 & - \\
 DMD (CFG=3)~\citep{yin2024one}  & 512 & 0.09s & 1 & 0.9B & 11.49 & -\\
  DMD (CFG=8)~\citep{yin2024one} & 512 & 0.09s & 1 & 0.9B & 14.98 & 0.320\\
  SD-Turbo~\citep{sdturbo} & 512 & 0.09s & 1 & 0.9B & 16.59 & 0.312\\
  \multicolumn{7}{c}{Distillation} \\
 BOOT~\citep{gu2023boot} & 512 & 0.09s & 1 & 0.9B & 48.20 & 0.26 \\
 Guided Distillation~\citep{guideddistillation} & 512 &  0.09s & 1 & 0.9B & 37.3 &0.27\\
 LCM~\citep{lcm} & 512 &  0.09s & 1 & 0.9B & 37.3 &0.27\\
 Phased Consistency Model~\citep{wang2024phased}& 512 &  0.09s & 1 & 0.9B & 17.91 &0.296\\
 Phased Consistency Model~\citep{wang2024phased}& 512 & 0.21s & 4 & 0.9B & 11.70 & - \\
  SiD-LSG ($\kappa=4.5$) &512& 0.09s & 1 & 0.9B & 16.59 & 0.317\\
   SiD-LSG ($\kappa=3$) &512& 0.09s & 1 & 0.9B & 13.21 & 0.314\\
  SiD-LSG ($\kappa=2$) &512& 0.09s & 1 & 0.9B & {9.56}&0.313\\
  SiD-LSG ($\kappa=1.5$) &512& 0.09s & 1 & 0.9B & 8.71&0.302\\

 SiD-LSG ($\kappa=4.5$) &512& 0.09s & 1 & 0.9B & 16.59 & 0.317\\
  \multicolumn{7}{c}{Rectification~($\star \star \star$)} \\
SDv1-5+DPMSolver~(Upper-Bound)~\citep{dpmsolver} & 512& 0.88s & 25 & 0.9B & 9.78 & 0.318\\\midrule
Rectified Flow~\citep{instaflow} & 512& 0.88s & 25 & 0.9B & 11.34 & 0.313\\ 
Rectified Flow~\citep{instaflow} & 512& 0.09s & 1 & 0.9B & 36.68 & 0.272\\ 
Rectified Flow~\citep{instaflow} & 512& 0.13s & 2 & 0.9B & 20.01 & 0.296\\ 
Rectified Diffusion~(Ours) & 512& 0.88s & 25 & 0.9B & 10.73 & 0.315\\ 
Rectified Diffusion~(Ours) & 512& 0.09s & 1 & 0.9B & 16.88 & 0.293\\ 
Rectified Diffusion~(Ours) & 512& 0.13s & 2 & 0.9B & 12.57 & 0.307\\ \midrule
Rectified Flow~(Distill)~\citep{instaflow}& 512& 0.09s & 1 & 0.9B & 13.67 & 0.302\\ 
Rectified Flow~(Distill)~\citep{instaflow}& 512& 0.13s & 2 & 0.9B & 62.81 & 0.261\\ 
Rectified Diffusion~(CD)~(Ours)& 512& 0.09s & 1 & 0.9B & 12.54 & 0.303\\ 
Rectified Diffusion~(CD)~(Ours)& 512& 0.13s & 2 & 0.9B & 11.41 & 0.310\\ \midrule
PeRFlow~\citep{yan2024perflow}& 512& 0.09s & 1 & 0.9B & 18.59 & 0.264\\ 
Rectified Diffusion~(Phased)~(Ours)& 512& 0.09s & 1 & 0.9B & 10.21 & 0.310\\ \midrule
    \multicolumn{7}{c}{Stable Diffusion XL~(3B) and its accelerated or distilled versions}\\ 
      \multicolumn{7}{c}{GANs}\\
      SDXL-Turbo~\cite{sdturbo} & 512& 0.15s & 1 & 3B & 24.57 & 0.337\\ 
      SDXL-Turbo~\cite{sdturbo} & 512& 0.34s & 4 & 3B & 23.19 & 0.334\\ 
      SDXL-Lightning~\citep{sdxl-lightning} & 1024& 0.35s & 1 & 3B & 23.92 & 0.316\\ 
      SDXL-Lightning~\citep{sdxl-lightning} & 1024& 0.71s & 4 & 3B & 24.56 & 0.323\\ 
      DMDv2~\citep{yin2024improved} & 1024& 0.35s & 1 & 3B & 19.01 & 0.336\\ 
      DMDv2~\citep{yin2024improved} & 1024& 0.71s & 4 & 3B & 19.32 & 0.332\\ 
      \multicolumn{7}{c}{Distillation}\\  
       LCM~\citep{lcm}& 1024& 0.35s & 1 & 3B & 81.62 & 0.275\\ 
       LCM~\citep{lcm}& 1024& 0.71s & 4 & 3B & 22.16 & 0.317\\ 
           Phased Consistency Model~\citep{wang2024phased}& 1024& 0.35s & 1 & 3B & 25.31 & 0.318\\ 
          Phased Consistency Model~\citep{wang2024phased}& 1024& 0.71s & 4 & 3B & 21.04 & 0.329\\ 
      \multicolumn{7}{c}{Rectification~($\star \star \star$)} \\
    PeRFlow-XL~\citep{yan2024perflow}& 1024& 0.71s & 4 & 3B & 20.99 & 0.334\\ 
Rectified Diffusion-XL~(Phased)~(Ours)& 1024& 0.71s & 4 & 3B & 19.71 & 0.340\\ 
 \bottomrule
\end{tabular}}
\captionsetup{labelformat=empty, labelsep=none, font=scriptsize}
\caption{Results of Stable Diffusion XL-based models are tested with COCO-2014 10k following the evaluation setting of DMDv2~\citep{yin2024improved}. Other results are tested with COCO-2014 30k following the karpathy split.}
\end{table}

\subsection{Comparison}

\noindent \textbf{Training cost.} 
Following the setup from the InstaFlow paper, we first use Stable Diffusion v1-5 and DPM-Solver~\citep{dpmsolver} to generate 1.6 million images. Since InstaFlow does not specify the prompts used, we generate images using a randomly sampled set of 1.6 million prompts. During the training of Rectified Diffusion, we used a batch size of 128 for a total of 200,000 iterations, resulting in a total of \(128 \times 200,000 = 25,600,000\) samples processed. In comparison, InstaFlow processed \(64 \times 70,000 + 1024 \times 25,000 = 30,080,000\) samples. Thus, our total training cost is lower than that of InstaFlow. Additionally, InstaFlow's total training time was 75.2 A100 GPU days, whereas our method required approximately 20 A800 GPU days. Typically, the training efficiency of an A800 is about 80\% of that of an A100. We attribute this significant reduction in training time to not using the LPIPS Loss~\citep{lipips}, which generally improves FID but incurs substantial memory and computational costs during the latent diffusion decoding process.
\textit{For the second-stage distillation,} we employ consistency distillation training with a batch size of 512 for 10,000 iterations, consuming a total of 4.6 A800 GPU days. In contrast, the distillation process described in the InstaFlow paper takes 110 A100 GPU days. Our training cost is approximately 3\% of the GPU days of InstaFlow's distillation process.

\noindent \textbf{Training speed.}
We monitor the performance of Rectified Diffusion in terms of FID and CLIP score at different stages of training. It was observed from Fig.~\ref{fig:training-iterations} that our method achieve superior one-step performance compared to Rectified Flow after just 20,000 iterations, with further significant improvements as training continued. At this stage, the number of samples processed was only about 8\% of the samples processed by Rectified Flow. This efficiency is largely because Rectified Diffusion does not require converting the original epsilon prediction diffusion model, which follows the DDPM form, into a v-prediction flow-matching model—a process that incurs significant computational cost. 

\noindent \textbf{Qualitative comparison.}
We present a comparison of the images generated by Rectified Diffusion and Rectified Flow across various scenarios in Fig.~\ref{fig:images-1} and Fig.~\ref{fig:images-2}. First, we can observe that the Rectified Flow model performs poorly at low step counts, producing only very blurry images in fewer than eight steps. Additionally, we notice that the images generated by PeRFlow are blurry and fail to reflect the content of the text. Moreover, the results generated by Rectified Flow~(Distill) remain relatively blurry and lack the ability for multi-step refinement, which limits its applicability. Rectified Diffusion shows clearly superiority in these settings.

\noindent \textbf{Quantitative comparison.} We calculate the FID~\citep{fid} and CLIP scores~\citep{clip} for different models on the COCO-2017 validation set~\citep{coco} and the 30k subset of the COCO-2014 validation set~\citep{coco}, respectively. As shown in Table~\ref{tab:coco-2017} and Table~\ref{tab:comparison}, our model consistently outperforms the methods based on rectified flow across both metrics, different scenarios, and various steps. It also achieves performance comparable to advanced distillation and GAN training methods.

\noindent \textbf{Human preference metrics.} To more comprehensively evaluate the model performance, we compare the outputs using human preference models. We follow the testing setup of Diffusion-DPO~\citep{wallace2024diffusion}, generating images with 500 unique prompts from the Pick-a-pic~\citep{kirstain2023pick} validation set for comparison. We used the Laion-Aesthetic Predictor~\citep{schuhmann2022laion}, Pickscore~\citep{kirstain2023pick}, HPSv2~\citep{wu2023human}, and ImageReward~\citep{xu2024imagereward} to score the generated results from each model individually and calculate the win rate of each model across these metrics. Our results, shown in Fig~\ref{fig:human}, consistently outperform the results of Rectified Flow-based models.

\noindent \textbf{CFG-influence.}
We show the performance comparison of FID and CLIP Score between Rectified Flow and Rectified Diffusion under different step counts and CFG values in Fig.~\ref{fig:cfg}. We observe that Rectified Diffusion consistently outperforms Rectified Flow, especially in the low-step regime. Additionally, we find that CFG has a significant impact on both Rectified Diffusion and Rectified Flow; even in the 1-step generation scenario, using an appropriate CFG value can still significantly enhance performance.

\section{Conclusion}
In conclusion, we rethink and investigate the essence of rectified flow. We demonstrate that retraining with pre-collected noise-image pairs is the most important factor. Building on this insight, we propose Rectified Diffusion, extending its scope to general diffusion forms. We identify that it is not straightness but first-order property is the essential training target of Rectified Diffusion. Additionally, by incorporating consistency distillation and introducing Rectified Diffusion~(Phased), we further enhance training efficiency and model performance, offering a streamlined approach to efficient high-fidelity visual generation. Vast validation demonstrates the  advancements of Rectified Diffusion.

%% file: secs/sec2_related_work.tex
\section{Related Works}

\noindent \textbf{Diffusion models.}  
Diffusion models have steadily become the foundational models in image synthesis~\citep{ddpm,sde,edm}. Extensive research has been conducted to explore their underlying principles~\citep{flowmathcing,riemannianflowmathcing,sde,vdm} and to expand or enhance the design space of these models~\citep{ddim,edm,vdm}. Additionally, several works have focused on innovating the model architecture~\citep{diffusionbeatgan,dit}, while others have scaled up diffusion models for text-conditioned image synthesis and various real-world applications~\citep{shi2024motion,sd,sdxl}. Moreover, efforts to accelerate sampling have been pursued at both the scheduler level~\citep{edm,dpmsolver,ddim} and the training level~\citep{guideddistillation,cm,zhou2024score,zhou2024long}. The former typically involves refining the approximation of the PF-ODE~\citep{dpmsolver,ddim}, while the latter focuses on distillation techniques~\citep{guideddistillation,progressivedistillation,cm,wang2024phased,wang2024animatelcm} or initializing diffusion weights for GAN training~\citep{sdturbo,sdxl-lightning,xu2024ufogen}.

\noindent \textbf{Rectified Flow.} \citet{flowmathcing} proposes the flow matching based on continuous normalizing flows, providing a different and unified perspective to understand diffusion models.  \citet{rectifiedflow} proposes the method rectified flow, setting up important baseline for diffusion acceleration and providing a solid theoretical analysis. It proposes rectification to straighten the ODE path of flow-matching based diffusion models. In the proof, \citet{rectifiedflow} show that the rectification allows for non-decreasing straightness of ODE. \citet{instaflow}  scale up the idea of rectified flow into large text-to-image generations, achieving one-step generation without introducing GAN.   \citet{yan2024perflow} proposes to split the ODE path into multi-phase following the InstaFlow~\citep{instaflow}. \citet{lee2024improving} analysises that one-time rectification is generally enough to achieve pure straightness and proposes better optimization strategy for enhanced performance of rectified flows.

\section{Limitations}
At low-step regime, the performance of methods based on rectification still lags behind state-of-the-art methods based on distillation~\citep{zhou2024score} or GAN training~\citep{yin2024improved,sdturbo}. Additional distillation steps are needed to improve low-step performance, which is also stated in InstaFlow~\citep{instaflow}.

%% file: secs/sec_theorem.tex
\section{Proof for first-order ODE}

\begin{theorem}\label{th:first-order}
 For the general diffusion form \(\mathbf x_t = \alpha_t \mathbf x_0 + \sigma_t \boldsymbol \epsilon\), there exists an exact ODE solution form as follows:
\begin{align}\label{eq:proof-exact}
    \mathbf x_t = \frac{\alpha_t}{\alpha_s} \mathbf x_s - \alpha_t \int_{\lambda_s}^{\lambda_t} e^{-\lambda} \boldsymbol \epsilon_{\boldsymbol \theta}(\mathbf x_{t_{\lambda}}, t_{\lambda}) \mathrm{d}\lambda,
\end{align}
where \(\lambda_t = \ln \frac{\alpha_t}{\sigma_t}\) and \(t_{\lambda}\) is the inverse function of \(\lambda_t\). The first-order ODE satisfies
\begin{align}\label{eq:proof-first-order}
    \mathbf x_t = \frac{\alpha_t}{\alpha_s} \mathbf x_s - \alpha_t \boldsymbol \epsilon_{\boldsymbol \theta}(\mathbf x_s, s) \int_{\lambda_s}^{\lambda_t} e^{-\lambda} \mathrm{d}\lambda = \frac{\alpha_t}{\alpha_s} \mathbf x_s - \alpha_t \boldsymbol \epsilon_{\boldsymbol \theta}(\mathbf x_s, s) (\frac{\alpha_s}{\sigma_s} - \frac{\alpha_t}{\sigma_t}) \, .
\end{align}
We show the equivalence between Equation~\ref{eq:proof-exact} and Equation~\ref{eq:proof-first-order} for arbitrary \(t\) and \(s\), which holds true \underline{if and only if} \(\boldsymbol \epsilon_{\boldsymbol \theta}(\mathbf x_t, t)\) is constant.
\end{theorem}

\begin{proofpart} 

\noindent \textbf{If \(\boldsymbol \epsilon_{\boldsymbol \theta}(\mathbf x_{t}, t)\) is constant, then the Equation~\ref{eq:proof-exact} and Equation~\ref{eq:proof-first-order} are equivalent.}

Assumption: Let \(\boldsymbol \epsilon_{\boldsymbol \theta}(\mathbf x_s, s) = \boldsymbol \epsilon_0\) be a constant.

Substituting $\boldsymbol \epsilon_0$ into the Equation~\ref{eq:proof-exact}:
   \begin{align}
   \mathbf x_t = \frac{\alpha_t}{\alpha_s} \mathbf x_s - \alpha_t \boldsymbol \epsilon_0 \int_{\lambda_s}^{\lambda_t} e^{-\lambda} \mathrm{d}\lambda
   \end{align}

Calculating the integral:
   \begin{align}
   \int_{\lambda_s}^{\lambda_t} e^{-\lambda} \mathrm{d}\lambda = e^{-\lambda_s} - e^{-\lambda_t} = \frac{\sigma_s}{\alpha_s} - \frac{\sigma_t}{\alpha_t}
   \end{align}

Substituting the result:
   \begin{align}
   \mathbf x_t = \frac{\alpha_t}{\alpha_s} \mathbf x_s - \alpha_t \boldsymbol \epsilon_0 \left(\frac{\sigma_s}{\alpha_s} - \frac{\sigma_t}{\alpha_t}\right)
   \end{align}

Comparing with the equation: The results match, thus proving equivalence.

\noindent \textbf{If Equation~\ref{eq:proof-exact} and Equation~\ref{eq:proof-first-order} are equivalent, then \(\boldsymbol \epsilon_{\boldsymbol \theta}(\mathbf x_{t}, t)\) must be constant.}

Assumption: Assume the two are equivalent:
   \begin{align}
   -\alpha_t \int_{\lambda_s}^{\lambda_t} e^{-\lambda} \boldsymbol \epsilon_{\boldsymbol \theta}(\mathbf x_{t_{\lambda}}, t_{\lambda}) \mathrm{d}\lambda = -\alpha_t \boldsymbol \epsilon_{\boldsymbol \theta}(\mathbf x_s, s) \int_{\lambda_s}^{\lambda_t} e^{-\lambda} \mathrm{d}\lambda
   \end{align}

Removing the constant factor:
   \begin{align}
   \int_{\lambda_s}^{\lambda_t} e^{-\lambda} \boldsymbol \epsilon_{\boldsymbol \theta}(\mathbf x_{t_{\lambda}}, t_{\lambda}) \mathrm{d}\lambda = \boldsymbol \epsilon_{\boldsymbol \theta}(\mathbf x_s, s) \int_{\lambda_s}^{\lambda_t} e^{-\lambda} \mathrm{d}\lambda
   \end{align}

Differentiating with respect to \(t\) with Newton-Leibniz theorem:
   \begin{align}
   \frac{d}{dt}\left(\int_{\lambda_s}^{\lambda_t} e^{-\lambda} \boldsymbol \epsilon_{\boldsymbol \theta}(\mathbf x_{t_{\lambda}}, t_{\lambda}) \mathrm{d}\lambda\right) = e^{-\lambda_t} \boldsymbol \epsilon_{\boldsymbol \theta}(\mathbf x_{t_{\lambda}}, t_{\lambda}) \frac{d\lambda_t}{dt}
   \end{align}

Comparing both sides:
   \begin{align}
   e^{-\lambda_t} \boldsymbol \epsilon_{\boldsymbol \theta}(\mathbf x_{t_{\lambda}}, t_{\lambda}) \frac{d\lambda_t}{dt} = \boldsymbol \epsilon_{\boldsymbol \theta}(\mathbf x_s, s) e^{-\lambda_t} \frac{d\lambda_t}{dt}
   \end{align}

   Since \(\frac{d\lambda_t}{dt} \neq 0\) and \(  e^{-\lambda_t} > 0\), we can cancel terms, leading to:
   \begin{align}
   \boldsymbol \epsilon_{\boldsymbol \theta}(\mathbf x_{t_{\lambda}}, t_{\lambda}) = \boldsymbol \epsilon_{\boldsymbol \theta}(\mathbf x_s, s), \forall t_{\lambda} \in [s, t] .
   \end{align}

Conclusion: This shows that for any \(t\), \(\boldsymbol \epsilon_{\boldsymbol \theta}(\mathbf x_t, t)\) must be constant, proving the ``if and only if" statement.
\end{proofpart}

\section{More results.}

\begin{figure}[t]
    \centering
    \includegraphics[width=1.0\linewidth]{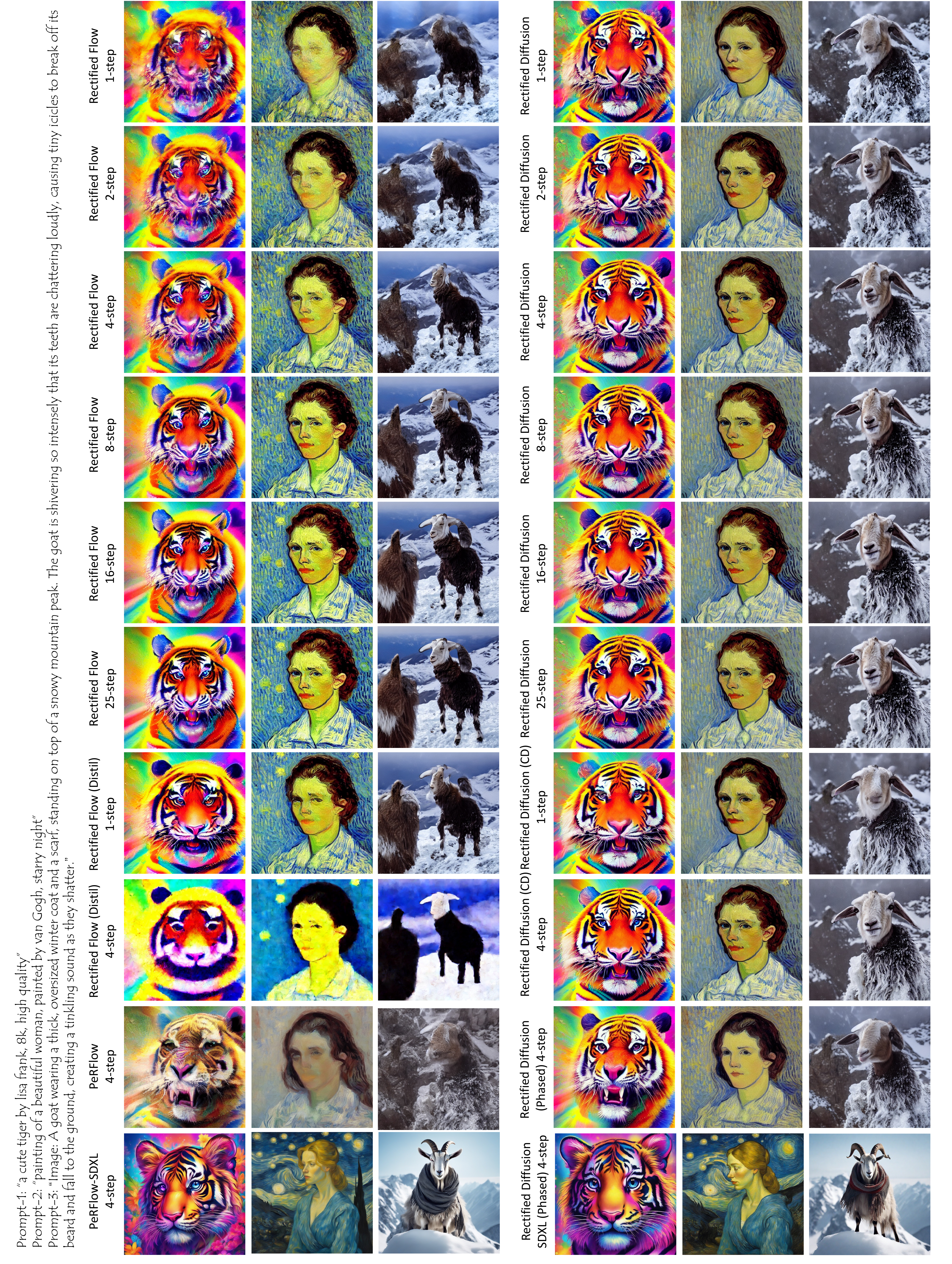}
    \caption{Qualitative comparison.}
    \label{fig:images-1}
\end{figure}

\begin{figure}[t]
    \centering
    \includegraphics[width=1.0\linewidth]{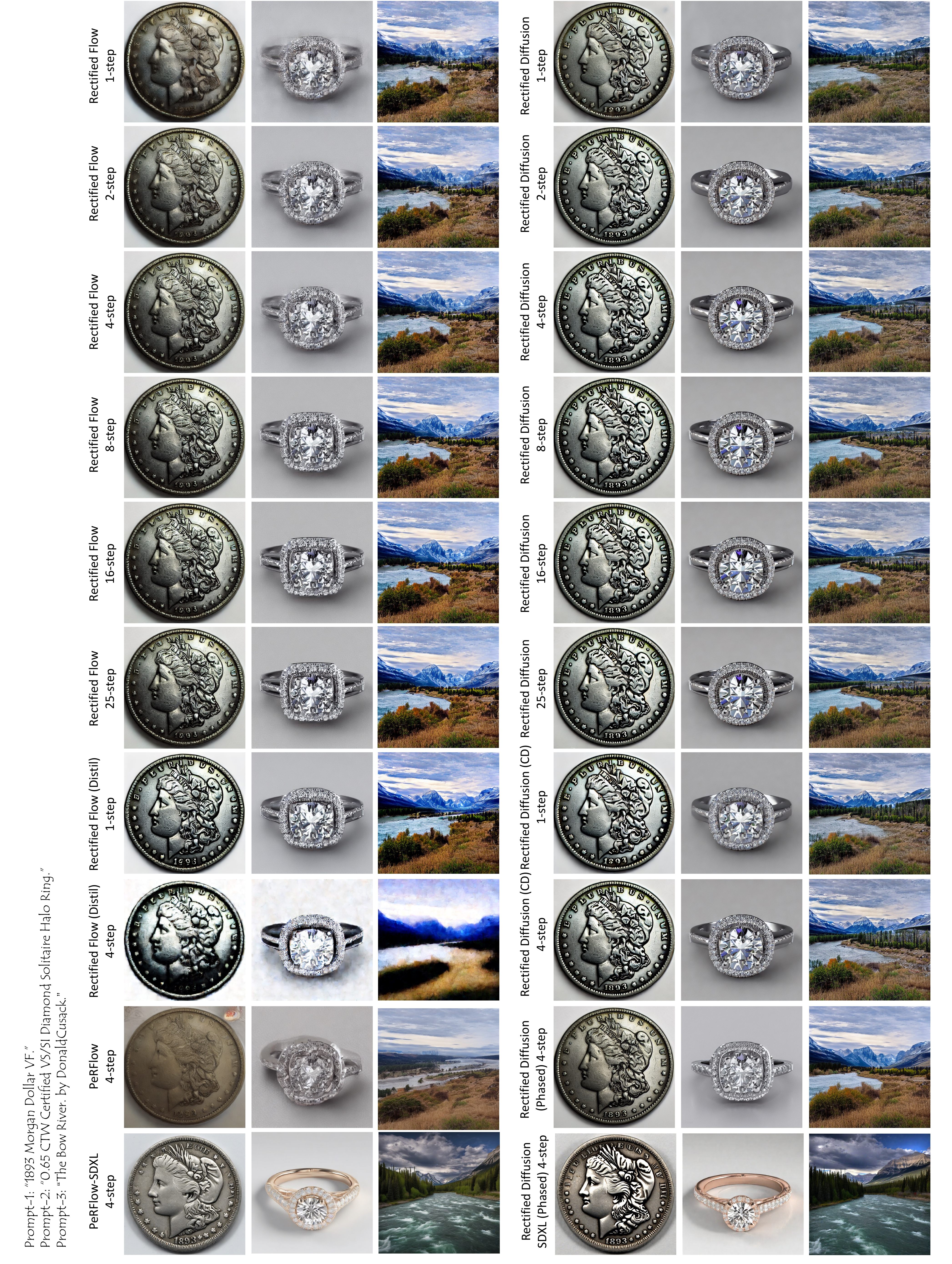}
    \caption{Qualitative comparison.}
    \label{fig:images-2}
\end{figure}

\begin{figure}[t]
    \centering
    \includegraphics[width=1.0\linewidth]{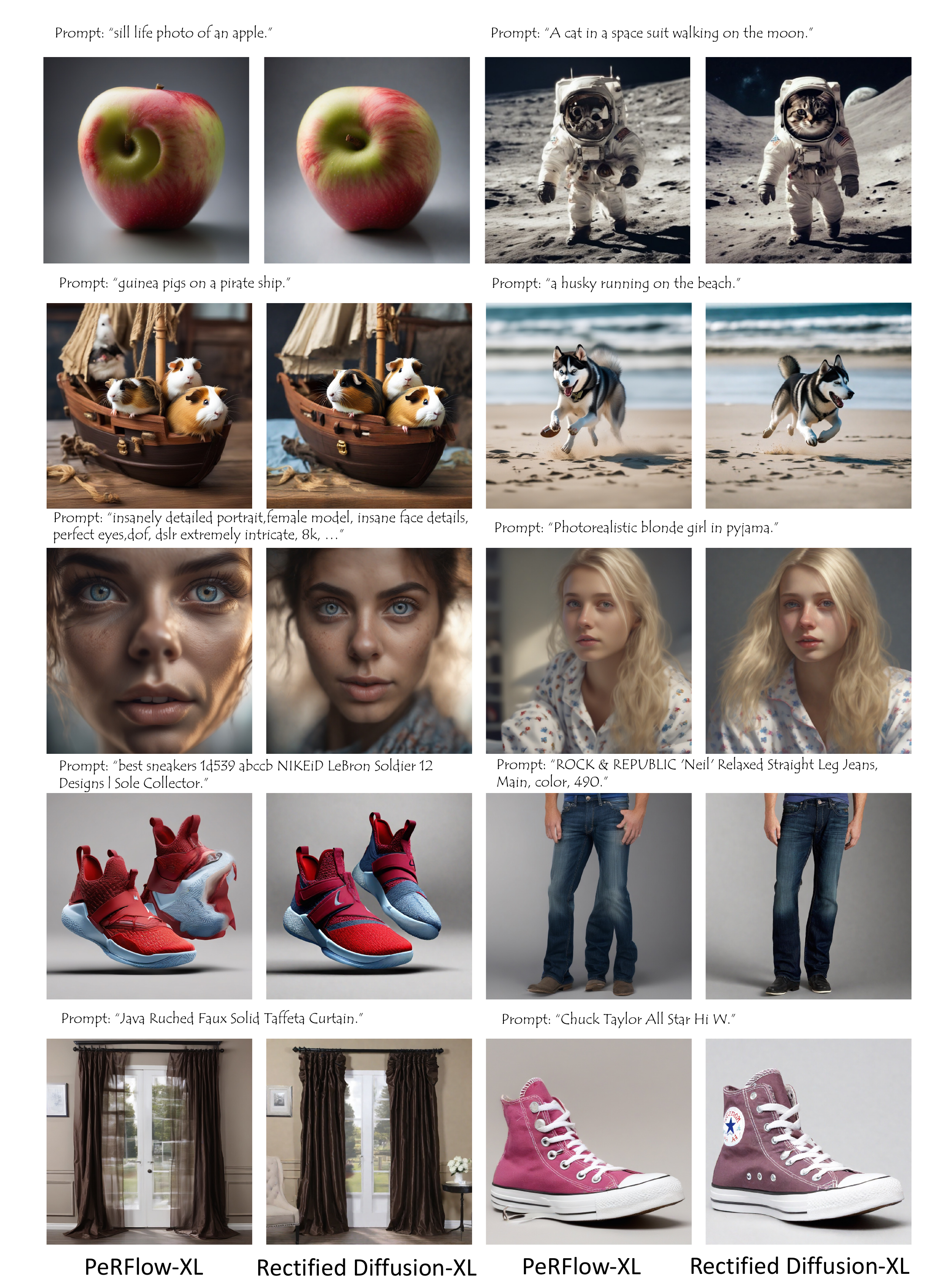}
    \caption{Qualitative comparison.}
    \label{fig:images-3}
\end{figure}

%% file: main.bbl
\begin{thebibliography}{57}
\providecommand{\natexlab}[1]{#1}
\providecommand{\url}[1]{\texttt{#1}}
\expandafter\ifx\csname urlstyle\endcsname\relax
  \providecommand{\doi}[1]{doi: #1}\else
  \providecommand{\doi}{doi: \begingroup \urlstyle{rm}\Url}\fi

\bibitem[Balaji et~al.(2022)Balaji, Nah, Huang, Vahdat, Song, Zhang, Kreis, Aittala, Aila, Laine, et~al.]{balaji2022ediffi}
Yogesh Balaji, Seungjun Nah, Xun Huang, Arash Vahdat, Jiaming Song, Qinsheng Zhang, Karsten Kreis, Miika Aittala, Timo Aila, Samuli Laine, et~al.
\newblock ediff-i: Text-to-image diffusion models with an ensemble of expert denoisers.
\newblock \emph{arXiv preprint arXiv:2211.01324}, 2022.

\bibitem[Chang et~al.(2023)Chang, Zhang, Barber, Maschinot, Lezama, Jiang, Yang, Murphy, Freeman, Rubinstein, Li, and Krishnan]{chang2023muse}
Huiwen Chang, Han Zhang, Jarred Barber, Aaron Maschinot, Jose Lezama, Lu~Jiang, Ming-Hsuan Yang, Kevin~Patrick Murphy, William~T. Freeman, Michael Rubinstein, Yuanzhen Li, and Dilip Krishnan.
\newblock Muse: Text-to-image generation via masked generative transformers.
\newblock In Andreas Krause, Emma Brunskill, Kyunghyun Cho, Barbara Engelhardt, Sivan Sabato, and Jonathan Scarlett (eds.), \emph{ICML}, volume 202 of \emph{Proceedings of Machine Learning Research}, pp.\  4055--4075. PMLR, 23--29 Jul 2023.
\newblock URL \url{https://proceedings.mlr.press/v202/chang23b.html}.

\bibitem[Chen \& Lipman(2023)Chen and Lipman]{riemannianflowmathcing}
Ricky~TQ Chen and Yaron Lipman.
\newblock Riemannian flow matching on general geometries.
\newblock \emph{arXiv preprint arXiv:2302.03660}, 2023.

\bibitem[Dhariwal \& Nichol(2021)Dhariwal and Nichol]{diffusionbeatgan}
Prafulla Dhariwal and Alexander Nichol.
\newblock Diffusion models beat gans on image synthesis.
\newblock \emph{NeurIPS}, 34:\penalty0 8780--8794, 2021.

\bibitem[Ding et~al.(2021)Ding, Yang, Hong, Zheng, Zhou, Yin, Lin, Zou, Shao, Yang, et~al.]{ding2021cogview}
Ming Ding, Zhuoyi Yang, Wenyi Hong, Wendi Zheng, Chang Zhou, Da~Yin, Junyang Lin, Xu~Zou, Zhou Shao, Hongxia Yang, et~al.
\newblock Cogview: Mastering text-to-image generation via transformers.
\newblock \emph{NeurIPS}, 34:\penalty0 19822--19835, 2021.

\bibitem[Esser et~al.(2024)Esser, Kulal, Blattmann, Entezari, M{\"u}ller, Saini, Levi, Lorenz, Sauer, Boesel, et~al.]{sd3}
Patrick Esser, Sumith Kulal, Andreas Blattmann, Rahim Entezari, Jonas M{\"u}ller, Harry Saini, Yam Levi, Dominik Lorenz, Axel Sauer, Frederic Boesel, et~al.
\newblock Scaling rectified flow transformers for high-resolution image synthesis.
\newblock \emph{arXiv preprint arXiv:2403.03206}, 2024.

\bibitem[Gafni et~al.(2022)Gafni, Polyak, Ashual, Sheynin, Parikh, and Taigman]{gafni2022make}
Oran Gafni, Adam Polyak, Oron Ashual, Shelly Sheynin, Devi Parikh, and Yaniv Taigman.
\newblock Make-a-scene: Scene-based text-to-image generation with human priors.
\newblock In \emph{ECCV}, pp.\  89--106. Springer, 2022.

\bibitem[Goodfellow et~al.(2020)Goodfellow, Pouget-Abadie, Mirza, Xu, Warde-Farley, Ozair, Courville, and Bengio]{gan}
Ian Goodfellow, Jean Pouget-Abadie, Mehdi Mirza, Bing Xu, David Warde-Farley, Sherjil Ozair, Aaron Courville, and Yoshua Bengio.
\newblock Generative adversarial networks.
\newblock \emph{Communications of the ACM}, 63\penalty0 (11):\penalty0 139--144, 2020.

\bibitem[Gu et~al.(2023)Gu, Zhai, Zhang, Liu, and Susskind]{gu2023boot}
Jiatao Gu, Shuangfei Zhai, Yizhe Zhang, Lingjie Liu, and Joshua~M Susskind.
\newblock Boot: Data-free distillation of denoising diffusion models with bootstrapping.
\newblock In \emph{ICML 2023 Workshop on Structured Probabilistic Inference $\{$$\backslash$\&$\}$ Generative Modeling}, 2023.

\bibitem[Heusel et~al.(2017)Heusel, Ramsauer, Unterthiner, Nessler, and Hochreiter]{fid}
Martin Heusel, Hubert Ramsauer, Thomas Unterthiner, Bernhard Nessler, and Sepp Hochreiter.
\newblock Gans trained by a two time-scale update rule converge to a local nash equilibrium.
\newblock \emph{NeurIPS}, 30, 2017.

\bibitem[Ho et~al.(2020)Ho, Jain, and Abbeel]{ddpm}
Jonathan Ho, Ajay Jain, and Pieter Abbeel.
\newblock ddpm.
\newblock \emph{NeurIPS}, 33:\penalty0 6840--6851, 2020.

\bibitem[Kang et~al.(2023)Kang, Zhu, Zhang, Park, Shechtman, Paris, and Park]{kang2023scaling}
Minguk Kang, Jun-Yan Zhu, Richard Zhang, Jaesik Park, Eli Shechtman, Sylvain Paris, and Taesung Park.
\newblock Scaling up gans for text-to-image synthesis.
\newblock In \emph{CVPR}, pp.\  10124--10134, 2023.

\bibitem[Karras et~al.(2022)Karras, Aittala, Aila, and Laine]{edm}
Tero Karras, Miika Aittala, Timo Aila, and Samuli Laine.
\newblock edm.
\newblock \emph{NeurIPS}, 35:\penalty0 26565--26577, 2022.

\bibitem[Kingma et~al.(2021)Kingma, Salimans, Poole, and Ho]{vdm}
Diederik~P Kingma, Tim Salimans, Ben Poole, and Jonathan Ho.
\newblock On density estimation with diffusion models.
\newblock In A.~Beygelzimer, Y.~Dauphin, P.~Liang, and J.~Wortman Vaughan (eds.), \emph{NeurIPS}, 2021.
\newblock URL \url{https://openreview.net/forum?id=2LdBqxc1Yv}.

\bibitem[Kirstain et~al.(2023)Kirstain, Polyak, Singer, Matiana, Penna, and Levy]{kirstain2023pick}
Yuval Kirstain, Adam Polyak, Uriel Singer, Shahbuland Matiana, Joe Penna, and Omer Levy.
\newblock Pick-a-pic: An open dataset of user preferences for text-to-image generation.
\newblock \emph{Advances in Neural Information Processing Systems}, 36:\penalty0 36652--36663, 2023.

\bibitem[Lee et~al.(2024)Lee, Lin, and Fanti]{lee2024improving}
Sangyun Lee, Zinan Lin, and Giulia Fanti.
\newblock Improving the training of rectified flows.
\newblock \emph{arXiv preprint arXiv:2405.20320}, 2024.

\bibitem[Lin et~al.(2024)Lin, Wang, and Yang]{sdxl-lightning}
Shanchuan Lin, Anran Wang, and Xiao Yang.
\newblock Sdxl-lightning: Progressive adversarial diffusion distillation.
\newblock \emph{arXiv preprint arXiv:2402.13929}, 2024.

\bibitem[Lin et~al.(2014)Lin, Maire, Belongie, Hays, Perona, Ramanan, Doll{\'{a}}r, and Zitnick]{coco}
Tsung{-}Yi Lin, Michael Maire, Serge~J. Belongie, James Hays, Pietro Perona, Deva Ramanan, Piotr Doll{\'{a}}r, and C.~Lawrence Zitnick.
\newblock Microsoft {COCO:} common objects in context.
\newblock In \emph{ECCV}, volume 8693, pp.\  740--755. Springer, 2014.

\bibitem[Lipman et~al.(2022)Lipman, Chen, Ben-Hamu, Nickel, and Le]{flowmathcing}
Yaron Lipman, Ricky~TQ Chen, Heli Ben-Hamu, Maximilian Nickel, and Matt Le.
\newblock Flow matching for generative modeling.
\newblock \emph{arXiv preprint arXiv:2210.02747}, 2022.

\bibitem[Liu et~al.(2022)Liu, Gong, and Liu]{rectifiedflow}
Xingchao Liu, Chengyue Gong, and Qiang Liu.
\newblock Flow straight and fast: Learning to generate and transfer data with rectified flow.
\newblock \emph{arXiv preprint arXiv:2209.03003}, 2022.

\bibitem[Liu et~al.(2023)Liu, Zhang, Ma, Peng, et~al.]{instaflow}
Xingchao Liu, Xiwen Zhang, Jianzhu Ma, Jian Peng, et~al.
\newblock Instaflow: One step is enough for high-quality diffusion-based text-to-image generation.
\newblock In \emph{ICLR}, 2023.

\bibitem[Lu et~al.(2022)Lu, Zhou, Bao, Chen, Li, and Zhu]{dpmsolver}
Cheng Lu, Yuhao Zhou, Fan Bao, Jianfei Chen, Chongxuan Li, and Jun Zhu.
\newblock Dpm-solver: A fast ode solver for diffusion probabilistic model sampling in around 10 steps.
\newblock \emph{NeurIPS}, 35:\penalty0 5775--5787, 2022.

\bibitem[Luo et~al.(2023)Luo, Tan, Huang, Li, and Zhao]{lcm}
Simian Luo, Yiqin Tan, Longbo Huang, Jian Li, and Hang Zhao.
\newblock Latent consistency models: Synthesizing high-resolution images with few-step inference.
\newblock \emph{arXiv preprint arXiv:2310.04378}, 2023.

\bibitem[Meng et~al.(2023)Meng, Rombach, Gao, Kingma, Ermon, Ho, and Salimans]{guideddistillation}
Chenlin Meng, Robin Rombach, Ruiqi Gao, Diederik Kingma, Stefano Ermon, Jonathan Ho, and Tim Salimans.
\newblock On distillation of guided diffusion models.
\newblock In \emph{CVPR}, pp.\  14297--14306, 2023.

\bibitem[Nichol et~al.(2021)Nichol, Dhariwal, Ramesh, Shyam, Mishkin, McGrew, Sutskever, and Chen]{nichol2021glide}
Alex Nichol, Prafulla Dhariwal, Aditya Ramesh, Pranav Shyam, Pamela Mishkin, Bob McGrew, Ilya Sutskever, and Mark Chen.
\newblock Glide: Towards photorealistic image generation and editing with text-guided diffusion models.
\newblock \emph{arXiv preprint arXiv:2112.10741}, 2021.

\bibitem[Peebles \& Xie(2023)Peebles and Xie]{dit}
William Peebles and Saining Xie.
\newblock Scalable diffusion models with transformers.
\newblock In \emph{ICCV}, pp.\  4195--4205, October 2023.

\bibitem[Podell et~al.(2023)Podell, English, Lacey, Blattmann, Dockhorn, M{\"u}ller, Penna, and Rombach]{sdxl}
Dustin Podell, Zion English, Kyle Lacey, Andreas Blattmann, Tim Dockhorn, Jonas M{\"u}ller, Joe Penna, and Robin Rombach.
\newblock Sdxl: Improving latent diffusion models for high-resolution image synthesis.
\newblock \emph{arXiv preprint arXiv:2307.01952}, 2023.

\bibitem[Radford et~al.(2021)Radford, Kim, Hallacy, Ramesh, Goh, Agarwal, Sastry, Askell, Mishkin, Clark, et~al.]{clip}
Alec Radford, Jong~Wook Kim, Chris Hallacy, Aditya Ramesh, Gabriel Goh, Sandhini Agarwal, Girish Sastry, Amanda Askell, Pamela Mishkin, Jack Clark, et~al.
\newblock Learning transferable visual models from natural language supervision.
\newblock In \emph{ICLR}, pp.\  8748--8763. PMLR, 2021.

\bibitem[Ramesh et~al.(2021)Ramesh, Pavlov, Goh, Gray, Voss, Radford, Chen, and Sutskever]{ramesh2021zero}
Aditya Ramesh, Mikhail Pavlov, Gabriel Goh, Scott Gray, Chelsea Voss, Alec Radford, Mark Chen, and Ilya Sutskever.
\newblock Zero-shot text-to-image generation, 2021.
\newblock URL \url{https://arxiv.org/abs/2102.12092}.

\bibitem[Ramesh et~al.(2022)Ramesh, Dhariwal, Nichol, Chu, and Chen]{ramesh2022hierarchical}
Aditya Ramesh, Prafulla Dhariwal, Alex Nichol, Casey Chu, and Mark Chen.
\newblock Hierarchical text-conditional image generation with clip latents.
\newblock \emph{arXiv preprint arXiv:2204.06125}, 1\penalty0 (2):\penalty0 3, 2022.

\bibitem[Rombach et~al.(2022{\natexlab{a}})Rombach, Blattmann, Lorenz, Esser, and Ommer]{rombach2022high}
Robin Rombach, Andreas Blattmann, Dominik Lorenz, Patrick Esser, and Bj{\"o}rn Ommer.
\newblock High-resolution image synthesis with latent diffusion models.
\newblock In \emph{CVPR}, pp.\  10684--10695, 2022{\natexlab{a}}.

\bibitem[Rombach et~al.(2022{\natexlab{b}})Rombach, Blattmann, Lorenz, Esser, and Ommer]{sd}
Robin Rombach, Andreas Blattmann, Dominik Lorenz, Patrick Esser, and Bj{\"o}rn Ommer.
\newblock High-resolution image synthesis with latent diffusion models.
\newblock In \emph{CVPR}, pp.\  10684--10695, 2022{\natexlab{b}}.

\bibitem[Saharia et~al.(2022)Saharia, Chan, Saxena, Li, Whang, Denton, Ghasemipour, Ayan, Mahdavi, Lopes, Salimans, Ho, Fleet, and Norouzi]{saharia2022photorealistic}
Chitwan Saharia, William Chan, Saurabh Saxena, Lala Li, Jay Whang, Emily Denton, Seyed Kamyar~Seyed Ghasemipour, Burcu~Karagol Ayan, S.~Sara Mahdavi, Rapha~Gontijo Lopes, Tim Salimans, Jonathan Ho, David~J Fleet, and Mohammad Norouzi.
\newblock Photorealistic text-to-image diffusion models with deep language understanding, 2022.
\newblock URL \url{https://arxiv.org/abs/2205.11487}.

\bibitem[Salimans \& Ho(2022)Salimans and Ho]{progressivedistillation}
Tim Salimans and Jonathan Ho.
\newblock Progressive distillation for fast sampling of diffusion models.
\newblock \emph{arXiv preprint arXiv:2202.00512}, 2022.

\bibitem[Sauer et~al.(2023{\natexlab{a}})Sauer, Karras, Laine, Geiger, and Aila]{sauer2023stylegan}
Axel Sauer, Tero Karras, Samuli Laine, Andreas Geiger, and Timo Aila.
\newblock Stylegan-t: Unlocking the power of gans for fast large-scale text-to-image synthesis.
\newblock In \emph{ICML}, pp.\  30105--30118. PMLR, 2023{\natexlab{a}}.

\bibitem[Sauer et~al.(2023{\natexlab{b}})Sauer, Karras, Laine, Geiger, and Aila]{stylegant}
Axel Sauer, Tero Karras, Samuli Laine, Andreas Geiger, and Timo Aila.
\newblock Stylegan-t: Unlocking the power of gans for fast large-scale text-to-image synthesis.
\newblock In \emph{ICLR}, pp.\  30105--30118. PMLR, 2023{\natexlab{b}}.

\bibitem[Sauer et~al.(2023{\natexlab{c}})Sauer, Lorenz, Blattmann, and Rombach]{sdturbo}
Axel Sauer, Dominik Lorenz, Andreas Blattmann, and Robin Rombach.
\newblock Adversarial diffusion distillation.
\newblock \emph{arXiv preprint arXiv:2311.17042}, 2023{\natexlab{c}}.

\bibitem[Schuhmann(2022)]{schuhmann2022laion}
Christoph Schuhmann.
\newblock Laion-aesthetics.
\newblock \url{https://laion.ai/blog/laion-aesthetics/}, 2022.
\newblock Accessed: 2023-11-10.

\bibitem[Shi et~al.(2024)Shi, Huang, Wang, Bian, Li, Zhang, Zhang, Cheung, See, Qin, et~al.]{shi2024motion}
Xiaoyu Shi, Zhaoyang Huang, Fu-Yun Wang, Weikang Bian, Dasong Li, Yi~Zhang, Manyuan Zhang, Ka~Chun Cheung, Simon See, Hongwei Qin, et~al.
\newblock Motion-i2v: Consistent and controllable image-to-video generation with explicit motion modeling.
\newblock In \emph{ACM SIGGRAPH 2024 Conference Papers}, pp.\  1--11, 2024.

\bibitem[Singer et~al.(2022)Singer, Polyak, Hayes, Yin, An, Zhang, Hu, Yang, Ashual, Gafni, et~al.]{makeavideo}
Uriel Singer, Adam Polyak, Thomas Hayes, Xi~Yin, Jie An, Songyang Zhang, Qiyuan Hu, Harry Yang, Oron Ashual, Oran Gafni, et~al.
\newblock Make-a-video: Text-to-video generation without text-video data.
\newblock \emph{arXiv preprint arXiv:2209.14792}, 2022.

\bibitem[Song et~al.(2020{\natexlab{a}})Song, Meng, and Ermon]{ddim}
Jiaming Song, Chenlin Meng, and Stefano Ermon.
\newblock Denoising diffusion implicit models.
\newblock \emph{arXiv preprint arXiv:2010.02502}, 2020{\natexlab{a}}.

\bibitem[Song et~al.(2020{\natexlab{b}})Song, Sohl-Dickstein, Kingma, Kumar, Ermon, and Poole]{sde}
Yang Song, Jascha Sohl-Dickstein, Diederik~P Kingma, Abhishek Kumar, Stefano Ermon, and Ben Poole.
\newblock Score-based generative modeling through stochastic differential equations.
\newblock \emph{arXiv preprint arXiv:2011.13456}, 2020{\natexlab{b}}.

\bibitem[Song et~al.(2023)Song, Dhariwal, Chen, and Sutskever]{cm}
Yang Song, Prafulla Dhariwal, Mark Chen, and Ilya Sutskever.
\newblock Consistency models.
\newblock \emph{arXiv preprint arXiv:2303.01469}, 2023.

\bibitem[Wallace et~al.(2024)Wallace, Dang, Rafailov, Zhou, Lou, Purushwalkam, Ermon, Xiong, Joty, and Naik]{wallace2024diffusion}
Bram Wallace, Meihua Dang, Rafael Rafailov, Linqi Zhou, Aaron Lou, Senthil Purushwalkam, Stefano Ermon, Caiming Xiong, Shafiq Joty, and Nikhil Naik.
\newblock Diffusion model alignment using direct preference optimization.
\newblock In \emph{CVPR}, pp.\  8228--8238, 2024.

\bibitem[Wang et~al.(2024{\natexlab{a}})Wang, Huang, Bergman, Shen, Gao, Lingelbach, Sun, Bian, Song, Liu, et~al.]{wang2024phased}
Fu-Yun Wang, Zhaoyang Huang, Alexander~William Bergman, Dazhong Shen, Peng Gao, Michael Lingelbach, Keqiang Sun, Weikang Bian, Guanglu Song, Yu~Liu, et~al.
\newblock Phased consistency model.
\newblock \emph{arXiv preprint arXiv:2405.18407}, 2024{\natexlab{a}}.

\bibitem[Wang et~al.(2024{\natexlab{b}})Wang, Huang, Shi, Bian, Song, Liu, and Li]{wang2024animatelcm}
Fu-Yun Wang, Zhaoyang Huang, Xiaoyu Shi, Weikang Bian, Guanglu Song, Yu~Liu, and Hongsheng Li.
\newblock Animatelcm: Accelerating the animation of personalized diffusion models and adapters with decoupled consistency learning.
\newblock \emph{arXiv preprint arXiv:2402.00769}, 2024{\natexlab{b}}.

\bibitem[Wu et~al.(2023)Wu, Hao, Sun, Chen, Zhu, Zhao, and Li]{wu2023human}
Xiaoshi Wu, Yiming Hao, Keqiang Sun, Yixiong Chen, Feng Zhu, Rui Zhao, and Hongsheng Li.
\newblock Human preference score v2: A solid benchmark for evaluating human preferences of text-to-image synthesis.
\newblock \emph{arXiv preprint arXiv:2306.09341}, 2023.

\bibitem[Xu et~al.(2024{\natexlab{a}})Xu, Liu, Wu, Tong, Li, Ding, Tang, and Dong]{xu2024imagereward}
Jiazheng Xu, Xiao Liu, Yuchen Wu, Yuxuan Tong, Qinkai Li, Ming Ding, Jie Tang, and Yuxiao Dong.
\newblock Imagereward: Learning and evaluating human preferences for text-to-image generation.
\newblock \emph{Advances in Neural Information Processing Systems}, 36, 2024{\natexlab{a}}.

\bibitem[Xu et~al.(2024{\natexlab{b}})Xu, Zhao, Xiao, and Hou]{xu2024ufogen}
Yanwu Xu, Yang Zhao, Zhisheng Xiao, and Tingbo Hou.
\newblock Ufogen: You forward once large scale text-to-image generation via diffusion gans.
\newblock In \emph{CVPR}, pp.\  8196--8206, 2024{\natexlab{b}}.

\bibitem[Yan et~al.(2024)Yan, Liu, Pan, Liew, Liu, and Feng]{yan2024perflow}
Hanshu Yan, Xingchao Liu, Jiachun Pan, Jun~Hao Liew, Qiang Liu, and Jiashi Feng.
\newblock Perflow: Piecewise rectified flow as universal plug-and-play accelerator.
\newblock \emph{arXiv preprint arXiv:2405.07510}, 2024.

\bibitem[Yin et~al.(2024{\natexlab{a}})Yin, Gharbi, Zhang, Shechtman, Durand, Freeman, and Park]{yin2024one}
Tianwei Yin, Micha{\"e}l Gharbi, Richard Zhang, Eli Shechtman, Fredo Durand, William~T Freeman, and Taesung Park.
\newblock One-step diffusion with distribution matching distillation.
\newblock In \emph{CVPR}, pp.\  6613--6623, 2024{\natexlab{a}}.

\bibitem[Yin et~al.(2024{\natexlab{b}})Yin, Gharbi, Park, Zhang, Shechtman, Durand, and Freeman]{yin2024improved}
Tianwei Yin, Michaël Gharbi, Taesung Park, Richard Zhang, Eli Shechtman, Fredo Durand, and William~T. Freeman.
\newblock Improved distribution matching distillation for fast image synthesis, 2024{\natexlab{b}}.
\newblock URL \url{https://arxiv.org/abs/2405.14867}.

\bibitem[Yu et~al.(2022)Yu, Xu, Koh, Luong, Baid, Wang, Vasudevan, Ku, Yang, Ayan, et~al.]{yu2022scaling}
Jiahui Yu, Yuanzhong Xu, Jing~Yu Koh, Thang Luong, Gunjan Baid, Zirui Wang, Vijay Vasudevan, Alexander Ku, Yinfei Yang, Burcu~Karagol Ayan, et~al.
\newblock Scaling autoregressive models for content-rich text-to-image generation.
\newblock \emph{arXiv preprint arXiv:2206.10789}, 2\penalty0 (3):\penalty0 5, 2022.

\bibitem[Zhang et~al.(2018)Zhang, Isola, Efros, Shechtman, and Wang]{lipips}
Richard Zhang, Phillip Isola, Alexei~A. Efros, Eli Shechtman, and Oliver Wang.
\newblock The unreasonable effectiveness of deep features as a perceptual metric.
\newblock In \emph{CVPR}, June 2018.

\bibitem[Zhou et~al.(2024{\natexlab{a}})Zhou, Wang, Zheng, and Huang]{zhou2024long}
Mingyuan Zhou, Zhendong Wang, Huangjie Zheng, and Hai Huang.
\newblock Long and short guidance in score identity distillation for one-step text-to-image generation.
\newblock \emph{arXiv preprint arXiv:2406.01561}, 2024{\natexlab{a}}.

\bibitem[Zhou et~al.(2024{\natexlab{b}})Zhou, Zheng, Wang, Yin, and Huang]{zhou2024score}
Mingyuan Zhou, Huangjie Zheng, Zhendong Wang, Mingzhang Yin, and Hai Huang.
\newblock Score identity distillation: Exponentially fast distillation of pretrained diffusion models for one-step generation.
\newblock In \emph{ICML}, 2024{\natexlab{b}}.

\bibitem[Zhou et~al.(2022)Zhou, Zhang, Chen, Li, Tensmeyer, Yu, Gu, Xu, and Sun]{zhou2022towards}
Yufan Zhou, Ruiyi Zhang, Changyou Chen, Chunyuan Li, Chris Tensmeyer, Tong Yu, Jiuxiang Gu, Jinhui Xu, and Tong Sun.
\newblock Towards language-free training for text-to-image generation.
\newblock In \emph{CVPR}, pp.\  17907--17917, 2022.

\end{thebibliography}
